\documentclass[12pt]{article}

\usepackage{amsmath,amsthm,amssymb}
\usepackage{commath}
\usepackage{xcolor}
\usepackage{algorithm}
\usepackage{algorithmicx}
\usepackage{algpseudocode}
\usepackage{subcaption}
\usepackage{fullpage}

\usepackage{scalerel,stackengine}
\usepackage{multirow}
\usepackage{cellspace}
\usepackage{setspace}
\usepackage{bm}
\usepackage{bbm}
\usepackage{graphicx}
\usepackage{tikz-cd,pgfplots}
\usepackage{framed}
\usepackage[utf8]{inputenc}
\usepackage[T1]{fontenc}
\usepackage[round]{natbib}
\usepackage{makecell}

\usepackage{hyperref}
\hypersetup{colorlinks,
            linkcolor=blue,
            citecolor=blue,
            urlcolor=magenta,
            linktocpage,
            plainpages=false}

\usepackage[capitalise]{cleveref}

\usepackage{varwidth}%

\DeclareMathOperator*{\argmin}{\arg\min}   

\newcommand{\w}{\mathbf{w}}
\newcommand{\z}{\mathbf{z}}
\newcommand{\x}{\mathbf{x}}
\newcommand{\M}{\mathbf{M}}
\newcommand{\y}{\mathbf{y}}

\newcommand{\g}{\mathbf{g}}
\newcommand{\xib}{\bm{\xi}}

\newcommand{\teta}{\tilde{\eta}}
\newcommand{\K}{\mathcal{K}}
\newcommand{\D}{\mathcal{D}}
\newcommand{\A}{\mathcal{A}}

\newcommand{\Expec}{\mathbb{E}}
\newcommand{\R}{\mathbb{R}}
\def\xbar{\bar{\mathbf{x}}}

\newcommand{\delaytaut}{{t-\tau_t}}
\newcommand{\delaytautone}{{t-1-\tau_{t-1}}}

\newcommand{\Reg}{\text{Reg}}
\newcommand{\bc}[1]{\left\{{#1}\right\}}
\newcommand{\br}[1]{\left({#1}\right)}

\newcommand{\MyNorm}[1]{\left\lVert#1\right\rVert}

\newcommand{\gradf}[1]{\nabla f({#1})}

\algdef{SE}[SUBALG]{Indent}{EndIndent}{}{\algorithmicend\ }%
\algtext*{Indent}
\algtext*{EndIndent}

\usepackage{authblk}
\newcounter{@inst}
\newcounter{@auth}

\newdimen\instindent
\newbox\authrun
\newtoks\authorrunning
\newtoks\tocauthor
\newbox\titrun
\newtoks\titlerunning
\newtoks\toctitle

\def\clearheadinfo{\gdef\@author{No Author Given}%
                  \gdef\@title{No Title Given}%
                  \gdef\@subtitle{}%
                  \gdef\@institute{No Institute Given}%
                  \gdef\@thanks{}%
                  \global\titlerunning={}\global\authorrunning={}%
                  \global\toctitle={}\global\tocauthor={}}

\stackMath
\newcommand\reallywidehat[1]{%
\savestack{\tmpbox}{\stretchto{%
  \scaleto{%
    \scalerel*[\widthof{\ensuremath{#1}}]{\kern.1pt\mathchar"0362\kern.1pt}%
    {\rule{0ex}{\textheight}}
  }{\textheight}%
}{2.4ex}}%
\stackon[-6.9pt]{#1}{\tmpbox}%
}

\theoremstyle{plain}
\newtheorem{theorem}{Theorem}
\newtheorem{lemma}[theorem]{Lemma}

\newtheorem{remark}[theorem]{Remark}
\theoremstyle{definition}
\newtheorem{definition}[theorem]{Definition}
\theoremstyle{definition}

\title{Learning Under Delayed Feedback: Implicitly Adapting to Gradient Delays}

\author[1]{Rotem Zamir Aviv}
\author[2]{Ido Hakimi}
\author[2]{Assaf Schuster}
\author[1,3]{Kfir Y. Levy}
\affil[1]{Department of Electrical and Computer Engineering, Technion}
\affil[2]{Department of Computer Science, Technion}
\affil[3]{A Viterbi Fellow}
\date{}

\begin{document}
\maketitle

\begin{abstract}
We consider stochastic convex optimization problems, where several machines act asynchronously in parallel while sharing a common memory. We propose a robust training method for the constrained setting and derive non asymptotic convergence guarantees that do not depend on prior knowledge of update delays, objective smoothness, and gradient variance. Conversely, existing methods for this setting crucially rely on this prior knowledge, which render them unsuitable for essentially all shared-resources computational environments, such as clouds and data centers. Concretely, existing approaches are unable to accommodate changes in the delays which result from dynamic allocation of the machines, while our method implicitly adapts to such changes.
\end{abstract}

\section{Introduction}
\label{sec:intro}
The past decade has witnessed a wide adoption  of machine learning (ML) techniques in various fields. However, the underlying model complexity and the vast amount of  data required to train modern ML models may lead to impractical prolonged training time. Parallelizing the learning process has the potential to greatly decrease the learning duration, by exploiting more computation power at every step.

Stochastic Gradient Descent (SGD) and its variants are amongst the most popular training methods in ML, and are known to work well even for modern large scale problems.
However, due to the sequential nature of SGD, extending its usage for distributed learning is not straightforward.

Distributed learning methods with parameter-server can be divided into synchronous Vs. asynchronous approaches.
In the synchronous setting, all of the machines  communicate at the same time and depend on one another to proceed. Such methods are easier to analyze, yet their performance depends on the slowest machine and they require large communication overheads.
Conversely, in the asynchronous case, machines may communicate independently of other machines, which allows more flexibility and reduces communication.  It is well known that the performance of asynchronous methods degrades due to the staleness of the gradient updates, i.e., \emph{the delay between the current model, and the (outdated) model for which the gradient feedback is computed}.

The asynchronous setting was extensively investigated in the context of stochastic convex optimization, which captures fundamental learning problems like linear regression, logistic regression and SVMs.
It was shown that if the learning objective is non-smooth, delays in the computed gradient feedback necessarily degrade the performance of SGD compared to the non-delayed setting, see e.g. \cite{joulani2013online, nedic2001distributed}.

Nevertheless, it was recently shown that the latter does not apply for smooth objectives.
\citet{arjevani2018tight} were the first to obtain the optimal rates for SGD with delays in the smooth convex case. They have provided a version of delayed SGD that incurs no degradation compared to the non-delayed case as long as  the maximal delay $\tau_{\text{max}}$ is small enough: $\tau_{\text{max}} \leq O(\sqrt{T})$, where $T$ is the total number of SGD updates. In the strongly-convex case they have shown that there is no degradation as long as   $\tau_{\text{max}} \leq O(T/\log T)$.
While the result of  \citet{arjevani2018tight} was limited to quadratic objectives, in a very recent work, 
\citet{stich2019error} have generalized their result for the general convex smooth case, while obtaining 
the same optimal guarantees.

Unfortunately, the optimal methods of \citet{arjevani2018tight} and \citet{stich2019error} \textbf{(i)} do not hold for constrained problems, \textbf{(ii)} degrade with the maximal delay, which might be substantially higher than the average delay, and are not able to accommodate changes in the delays throughout the training process, and \textbf{(iii)} require prior knowledge of the maximal delay and smoothness parameter. In practice, when working with shared resources computational systems, delays might vary with time due to dynamic allocation of machines. Such scalable systems are central to the entire discipline of distributed learning, when more and more computations are carried on remote machines. 

Thus, we propose delay-adaptive and delay-robust algorithms for asynchronous stochastic convex optimization, with arbitrary time varying delays. \emph{Delay-robust} means that one can tune a baseline (online) algorithm for a given model without delays, yet apply it for a delay incorporated model, without further tuning, while achieving the optimal delay dependency (or vice versa). By \emph{delay-adaptive} we mean that our methods do not require any information regarding the delays, and are able to accommodate non-stationary changes in the delays.

\textbf{Contributions:} We summarize our contributions below,
\begin{itemize}
    \item For the general convex smooth case we utilize a simple yet general approach to develop a \emph{delay-adaptive} algorithm that obtains the optimal rates for the \emph{delayed constrained} setting, thus resolving an open question posed by \citet{stich2019error}. Our algorithm implicitly adapts to the objective smoothness and the gradient variance.
    \item For the strongly-convex smooth case we develop a \emph{delay-adaptive} algorithm which obtains meaningful yet suboptimal guarantees for the delayed constrained setting, without prior knowledge of the smoothness parameter or the gradient variance. This is the first method for the delayed strongly-convex and \emph{constrained} setting with non-trivial guarantees.
    \item We allow for arbitrary delays that may vary with  time, with no further assumptions. Our algorithm implicitly adapts to changes in delays, which enhances its robustness and enables usage in scalable systems.  In contrast to previous works on this topic, the performance of our method degrades \emph{proportionally to the average delay rather than to the maximal one.}
    \item We validate the performance of our algorithm on real-world data. The experiments demonstrate our algorithm robustness and adaptivity. Concretely, 
    when tuning SGD to train a given model for a specific delay regime and then changing the delay regime, its performance might degrade.
    Conversely, our algorithm maintains its high performance.
\end{itemize}
On the technical side, our work builds on a recent online to batch technique \cite{cutkosky2019anytime, kavis2019unixgrad}, that we combine with optimistic online learning techniques \cite{mohri2016accelerating, rakhlin2013optimization}.

\subsection{Related Work}
There is a large volume of published studies considering distributed learning; we provide few closely related examples. A recent survey on this topic can be viewed e.g. in \cite{verbraeken2020survey,ben2019demystifying}.

Our focus here is on centralized distributed setting, where there is a single parameter-vector that is updated using information received from several parallel machines.  
The centralized case further divides into synchronous Vs. asynchronous approaches.
Synchronous training methods usually employ large batches to compute  gradient estimates, when the batch computation is distributed between the machines. This approach was extensively investigated e.g.~in
\cite{dekel2012optimal,cotter2011better,shalev2013accelerated,li2014efficient,takavc2015distributed,jain2016parallelizing}.

The centralized asynchronous case was 
investigated e.g. in  \cite{bertsekas1989parallel,agarwal2012distributed,shamir2014distributed,mcmahan2014delay,sra2015adadelay,dutta2018slow}. 
One line of work that has gained much interest in the asynchronous case is  a model where the updates in the parameter-vector are performed with individual coordinate granularity  \cite{recht2011hogwild,leblond2018improved}. It was shown that this approach is beneficial when data features are sparse.

Another line of work in the context of asynchronous training is to analyze SGD with delayed gradient feedbacks. This study was initiated by \cite{agarwal2012distributed}, and followed by 
\cite{mcmahan2014delay,sra2015adadelay,lian2015asynchronous,feyzmahdavian2016asynchronous,zheng2017asynchronous,dutta2018slow,arjevani2018tight,stich2019error} amongst others.
 
An optimal SGD variant for the delayed convex smooth case was first suggested by \citep{arjevani2018tight}, providing an optimal convergence rate  of $O(1/\sqrt{T}+\tau/T)$, where 
$\tau$ is some constant delay and $T$ is the total number of gradient updates. They also proved an optimal rate of $O\left(1/T + \exp{\left(\frac{-HT}{L\tau}\right)}\right)$ for the strongly-convex and smooth case, where $L$ and $H$ are the smoothness and strong-convexity parameters of the objective.
These bounds imply that we suffer no degradation compared to  non-delayed SGD, as long as $\tau \leq O(\sqrt{T})$ in the convex case, and $\tau \leq O(T/\log{T})$ in the strongly-convex case.
While the method of \citep{arjevani2018tight} only applies for quadratic losses, the very recent work of  \cite{stich2019error} has generalized these results to the general convex case and upper bounded delays $\tau \in[0,\tau_{\max}]$.

Finally, while previous papers assume upper bounded delays \cite{lian2015asynchronous}, \cite{agarwal2012distributed}, \cite{zheng2017asynchronous}, in environments such as clouds and data centers this assumptions is often violated. To ensure scalability, the algorithm must accommodate changes in delays. \citet{sra2015adadelay,mcmahan2014delay,dutta2018slow,ren2020delay,zhou2018distributed} have addressed this issue, though with either some additional assumptions on the delay distribution, suboptimal guarantees, or asymptotic guarantees.

The rest of the paper is organized as follows. The next section describes our problem formulation and introduces the concept of Online Convex Optimization. Section \ref{sec:AnytimeAll} offers our main Theorem for smooth non strongly convex objectives. Section \ref{sec:strongConv} presents the main Theorem for smooth strongly convex objectives, and in Section \ref{sec:exp} we detail our experimental setup and results. Complete proofs can be found in the supplement.

\section{Problem Formulation}
\label{sec:Setting}
We focus on solving the following optimization problem,
\[\min_{\w \in \K} f(\w)~,\]
where $f:\K \to \R$ is a convex and smooth function, and $\K$ is a compact convex subset of $\R^d$. 

We consider first-order iterative optimization algorithms with access to a stochastic gradient oracle that returns unbiased estimates of the objective gradients. In the standard non-delayed setup, in each iteration $t$ the algorithm queries a noisy gradient oracle with a point $\x_t\in\K$, and in return it receives an unbiased estimate, \[\g_t = \nabla f(\x_t) + \xib_t\;\;\forall t~,\] where $\Expec[\xib_t \vert\x_t]=0$. After $T$ iterations the  algorithm outputs an estimate $\x_T$ of the optimal solution, and its accuracy is measured by the expected excess loss,
\[\Expec[f(\x_T)-f(\w^*)]~,\] 
where $\w^*\in\argmin_{\w\in\K}f(\w)$.

In this paper we consider centralized distributed asynchronous learning problems. Concretely, we assume a central server which maintains a global parameter-vector, and employs several workers in parallel to update that vector using first order information. Each worker queries the central server for the most updated parameter-vector, and computes its gradients with respect to that vector. The gradients are sent back to the central server, which then updates its parameter-vector accordingly. Since every worker communicates with the central server independently of the others, by the time a certain worker has restored the gradients to the server, another may have updated the parameter-vector. Thus, the server updates the model based on stale gradients.

 This setting can be described as a first order stochastic optimization problem, yet with a  \emph{delayed noisy gradient oracle}. 
Now, when we query this oracle with $\x_t$, a stale gradient estimate of the objective at a previous query point $\x_\delaytaut$ is received. We assume that the delays $\tau_t \in[0,t-1]$ may change arbitrarily  and are unknown in advance. Similarly to \cite{arjevani2018tight}, we describe the \emph{delayed  noisy gradient oracle} as follows,
\[\g_\delaytaut = \nabla f(\x_\delaytaut) + \xib_t~,\]
where $\Expec[\xib_t \vert\x_t]=0$. In Appendix~\ref{sec:Noise} we explain why  $\Expec[\xib_t \vert\x_t]=0$  makes sense in the context of the asynchronous distributed setting. Note that in order for this assumption to hold for asynchronous stochastic optimization, all workers must have equal access to the data and the delays are assumed to be data-independent.

We make the following standard assumptions throughout the paper:
\begin{enumerate}
    \item Bounded gradients. There exists a constant $G>0$ such that \[\Vert\nabla f(\x)\Vert \leq G \;\;\; \forall \x\in\K~.\] 
    \item Bounded variance. There exist a constant $\sigma^2$ such that \[\Expec[\Vert\xib_t\Vert^2| \x_t]\leq\sigma^2\;\;\;\forall t~.\]
    \item Compact domain. There exist a constant $D$ such that \[\Vert\x-\y\Vert^2\leq D^2\;\;\;\forall \x,\y\in\K~.\]
    \item We assume $\tau_t\leq t-1$, since the first computed gradient is $\g_1$.
\end{enumerate}

\textbf{Online Convex Optimization.} 
Our results rely on Online Convex Optimization (OCO)  techniques that we use as a mechanism for solving the aforementioned stochastic optimization problem.  Next, we describe this setting and necessary definitions.

OCO problems can be depicted as a repeated game of $T$ rounds.
At every round $t\in[T]$ a learner makes a prediction $\w_t \in \K$, after which a convex loss function $f_t :\K \mapsto \R$ is chosen. Then, the learner incurs a loss $f_t(\w_t)$, and receives $f_t(\cdot)$ as a feedback.
We assume that the losses $f_t(\cdot)$ may change arbitrarily, and may depend on the choices of the learner up to round $t$. Now, given a sequence of non-negative weights $\{\alpha_t\}_{t\in[T]}$, the goal of the learner is to minimize the (weighted) regret, which is defined below,
\begin{equation}
    \Reg_T(\w^*) = \sum_{t=1}^T \alpha_t f_t(\w_t) - \min_{\w\in \K} \sum_{t=1}^T \alpha_t f_t(\w)~.
\end{equation}
Note that if $\alpha_t=1 \forall t$ this is the standard definition of regret. Oftentimes we assume that the learner can access $f_t(\cdot)$ through a first order oracle, i.e., he may query a gradient oracle of $f_t(\cdot)$.

There is a strong connection between OCO and stochastic optimization.
Concretely, if the losses $\{ f_t(\cdot)\}_{t\in[T]}$ received by the OCO algorithm are unbiased estimates of 
a fixed function $f(\cdot)$, i.e., $f_t(\w) := f(\w;z_t)$ where $z_t$'s are i.i.d. samples from some unknown distribution $\D$, then one can show the following online to batch conversion between regret guarantees and excess loss  \cite{cesa2004generalization},
\[\Expec\left[f\left(\frac{\sum_{t=1}^T \alpha_t\w_t}{\sum_{t=1}^T\alpha_t}\right)-f(\w^*)\right]\leq \frac{\Reg_T(\w^*)}{\sum_{t=1}^T\alpha_t}~,\]
where $f(\w): = \Expec_{z\sim \D}  [f(\w;z)]$.
This conversion allows usage of OCO methods for offline stochastic optimization, while maintaining the powerful framework of OCO and its strong guarantees.
The next section shows how our results build on a recent novel online to batch conversion that is different than the standard one that we have just described.

\paragraph{Preliminaries}

\begin{definition}
\textit{Smoothness.} Function $f$ is said to be smooth if and only if its gradient is $L$ Lipschitz with some $L>0$ over $\K$. Meaning,
\[\Vert \nabla f(\x) - \nabla f(\y)\Vert \leq L \Vert \x-\y\Vert\;\;\forall \x,\y \in \K~.\] 
\end{definition}

\begin{definition}
\textit{Strong convexity.} Function $f$ is said to be $H$ strongly-convex if and only if
\[f(\x)-f(\y) \leq \nabla f(\x)^\top(\x-\y) - \frac{H}{2}\Vert \x-\y\Vert^2\;\;\forall \x,\y \in \K~.\] 
\end{definition}

\begin{definition}
\textit{Projection Operation.} Let $\Pi_\K(\cdot)$ denote the projection operation onto set $\K$. Namely,
\[\Pi_\K(\y) \triangleq \argmin_{\x \in \K} \Vert \x-\y\Vert~.\] 
\end{definition}

\textbf{Notations.} We denote $\alpha_{1:t}=\sum_{i=1}^t \alpha_i$,  $\|\cdot\|$ as the Euclidean norm,  and $[t]:=\{1,\ldots,t\}$.
For a series of delays $\tau_1, \cdots, \tau_T$ we denote their average by $\mu_\tau$ and their variance by $\sigma_\tau$, i.e.,
\begin{align*}
    \mu_\tau: = \frac{1}{T}\sum_{t=1}^T \tau_t~~~~\&~~~~\sigma_\tau^2 := \frac{1}{T}\sum_{t=1}^T \tau_t^2-\mu_\tau^2~.
\end{align*}


\section{Delay Adaptive Scheme for General Convex Case}
\label{sec:AnytimeAll}

In this section we suggest two algorithms for the  stochastic delayed setting, assuming that
the objective $f(\cdot)$ is convex and smooth. In Section~ \ref{sec:GeneralOL} we analyze a general scheme that enables to take any OCO algorithm and turn it to a delay-adaptive and delay-robust stochastic optimization algorithm.
This scheme \emph{does not require any prior knowledge of the delays} or their statistics, and furthermore it \emph{does not require any tuning that depends on the delays}.

In Section~\ref{sec:optimistism} we describe and analyze a fully adaptive algorithm for the 
delayed setting. This algorithm obtains the optimal guarantees for this setting, while requiring the knowledge of neither the delays, nor any other problem parameters like noise variance, smoothness and gradient scale.

The difficulty of learning in the delayed setting stems from the difference between the stale gradient that we receive and the true gradient of the current iterate. This difference can be related directly to the degradation in the performance, where larger differences  lead to worse performance.
In the smooth case one can relate the difference between gradients to the difference between query points.
Examining this difference for standard SGD with a fixed learning rate $\eta$ shows that the distance between $\w_t$ and $\w_{t-\tau_t}$ can be bounded by $O(\tau_t \eta G)$ (using the update rule of SGD). Nevertheless, (optimizing over $\eta$) this bound is  of the order of $O(G\sqrt{\tau_t /T})$, which leads to suboptimal performance. 

Thus, in the context of the delayed setting, we would wish for an algorithm that employs \emph{slowly changing query points}. Ideally, we would like to employ queries $\{\x_t\}_{t\in[T]}$ such that 
$\|\x_t -\x_{t-\tau_t}\| \leq O(\tau_t/t)$, which is significantly smaller than the  $O(\sqrt{\tau_t /T})$ that we have seen for standard SGD.
Fortunately, it was recently demonstrated in \cite{cutkosky2019anytime} and \cite{kavis2019unixgrad} that it is possible to achieve slowly varying queries, while still maintain same guarantees as of standard SGD. They proposed an alternative approach to online to batch conversion, which queries the gradients at the \emph{iterate averages} rather than the iterates themselves. When the objective is smooth, the use of such a conversion scheme stabilizes the predictions of the algorithm and thus leads to optimal  convergence rates for the delayed setting. In other words, by querying the gradient oracle at the running averages, we implicitly adapt to gradient delays. We follow \cite{cutkosky2019anytime, kavis2019unixgrad}, and relate to this scheme as \emph{anytime online to batch conversion}.

\subsection{General Delay-Adaptive Scheme}
\label{sec:GeneralOL}
In this section we analyze algorithm \ref{Alg:AnytimeTaut}, which utilizes a general OCO algorithm $\A$, together with anytime online to batch conversion. Recall that in the standard online to batch scheme, gradients are queried at the iterates of the OCO algorithm.
Conversely, in the anytime online to batch scheme that we utilize in algorithm~\ref{Alg:AnytimeTaut}, the OCO algorithm $\A$ produces iterates $\w_t$, while the gradients that $\A$ receives are queried at $\x_t$'s, which are weighted averages of the iterates. 

By querying the gradients at the average we ensure slowly varying query points, even when the OCO iterates change rapidly. This allows us to receive general convergence guarantees for the delayed setting, compatible with any number of OCO algorithms, including such which are not tuned for the delayed setting. The result is stated in Theorem \ref{theo:GeneralAnytimeDelayTaut}.

\begin{algorithm}[t]
\caption{Delay Adaptive Anytime Online to Batch}
 \label{Alg:AnytimeTaut}
 \begin{algorithmic}
    \State {\bfseries Input:} \# of iterations $T$ , $\w_1 \in \K$,  weights $\bc{\alpha_t}_{t \in [T]}$
    \For{$t=1$ {\bfseries to} $T$}
   \State $\x_t \gets \frac{\sum_{i=1}^t\alpha_i\w_{i}}{\alpha_{1:t}}$
    \State get $\g_\delaytaut$ from worker
    \State  define $f_t(\x)=\g_\delaytaut^\top \x$
    \State send $\alpha_t f_t(\x)$ to $\mathcal{A}$
    \State get $\w_{t+1}$ from $\mathcal{A}$
   \EndFor
    \State \textbf{return} $\x_T$
 \end{algorithmic}

\end{algorithm}

\begin{theorem}
Assume that $f:\K\mapsto \R$ is $L$-smooth. Let $\Reg_T(\w^*)$ be the regret of $\mathcal{A}$ with respect to the following sequence 
$\{f_t(\x):=\g_\delaytaut^\top \x\}_{t\in[T]}$,
\begin{equation}
    \Reg_T(\w^*) := \sum_{t=1}^T\alpha_t\g_\delaytaut^\top(\w_t-\w^*)~.
    \label{eq:BoundGeneralRT}
\end{equation}

Then, using Alg.~\ref{Alg:AnytimeTaut} with $\alpha_t=t$ ensures,
\[\Expec\left[f(\x_T)-f(\w^*)\right] = O\left(\frac{\Reg_T(\w^*)}{T^2} + LD^2\frac{\mu_\tau}{T}\right)~,\]
where $\mu_\tau$ is the average delay.
\label{theo:GeneralAnytimeDelayTaut}
\end{theorem}

Theorem \ref{theo:GeneralAnytimeDelayTaut} implies that by using anytime online to batch conversion it is possible to convert any OCO algorithm into a delay-robust and delay-adaptive algorithm. In addition, while previous works obtained error bounds which are proportional to $O(\tau_{\max}/T)$, our bounds depend on $O(\mu_{\tau}/T)$ which might be substantially smaller.

 Note that there are  several standard algorithms that can be plugged into Alg. \ref{Alg:AnytimeTaut}, such as OGD (Online Gradient Descent), and FTRL (Follow the Regularized Leader) \cite{zinkevich2003online}, \cite{shalev2011online}. 
For example, a delay adaptive version of SGD, can be applied by using OGD as  $\A$ inside Alg.~\ref{Alg:AnytimeTaut}. The update rule in this case boils down to,
\begin{align}\label{eq:SGDaDAPTIVEdEL}
\w_{t+1} = \Pi_\K(\w_t-\teta_t\g_\delaytaut)~.
\end{align}
where $\teta_t \propto  \alpha_t/\sqrt{\sum_{i=1}^t\alpha_i^2} \approx 1/\sqrt{t}$, and  $\g_{t-\tau_t}$ is a stale gradient of one of the past query points. Moreover, the query points $\{\x_t\}_{t\in[T]}$ are \emph{weighted averages of past iterates; i.e., $\x_t = \sum_{i=1}^t \w_i/\alpha_{1:t}$}. 
Note that in this case, $\teta_t$ does not depend on the delays.

The average (weighted) regret of OGD can be shown to be,
\[O\left(\frac{\Reg_T(\w^*)}{\alpha_{1:T}} \right)=O({GD}/{\sqrt{T}})~.\]
Combining this with Theorem~\ref{theo:GeneralAnytimeDelayTaut} implies that delay adaptive SGD (Eq.~\eqref{eq:SGDaDAPTIVEdEL}) obtains an overall rate of
$
O(GD/\sqrt{T} + LD^2\mu_\tau/T)~.
$
For completeness, we include in Section \ref{sec:DelayedSGD} a proof for the OGD average regret bound. 
The full proof of Theorem \ref{theo:GeneralAnytimeDelayTaut} can be seen in the supplementary material. We present here a rough description of it.
\begin{proof}[Proof Sketch] First, using the gradient inequality together with $ \alpha_t(\x_t-\w_t)=\alpha_{1:t-1}(\x_{t-1}-\x_t)$, which follows from the definition of $\x_t$, we can decompose as follows,
\begin{align}
    \Expec\left[\sum_{t=1}^T \alpha_t(f(\x_t)-f(\w^*))\right] 
    \leq& \Expec\left[\sum_{t=1}^T \alpha_{1:t-1}\nabla f(\x_t)^\top(\x_{t-1}-\x_t)+\Reg_T(\w^*)\right] \nonumber\\
    &+\Expec\left[\sum_{t=1}^T\alpha_t\Vert\nabla f(\x_t)-\nabla f(\x_\delaytaut)\Vert\Vert\w_t-\w^*\Vert\right]~,
    \label{eq:ExpeLossMainBody}
\end{align}
where $\Reg_T(\w^*)$ is the  regret of $\A$ with respect to the sequence $\{f_t(\x): = \alpha_t\g_{t-\tau_t}^\top \x\}_{t\in[T]}$.
Due to the smoothness of the objective, bounding the last term is equivalent to bounding $\Vert \x_t - \x_\delaytaut\Vert$. Since $\x_t$ is a weighted average of the $\w_t$'s, one can show that,
\begin{equation}
    \Vert \x_t - \x_\delaytaut\Vert = O\left( {\tau_tD}/{t}\right)~.
    \label{eq:BoundXMainBody}
\end{equation}

Note that querying the oracle at the iterate averages rather than the iterates themselves was a key factor for establishing this bound. Using Eq. \eqref{eq:BoundXMainBody}, the last term of Eq. \eqref{eq:ExpeLossMainBody} is bounded by $LD^2\mu_\tau T$. From here we follow similar steps as in \cite{cutkosky2019anytime} and show that,
$$
\alpha_{1:T} \Expec\left[(f(\x_T)-f(\w^*))\right] \leq \Reg_T(\w^*) + O\left(\sum_{t=1}^T \alpha_t \frac{\tau_t D^2}{t}\right)~.
$$
Plugging $\alpha_t=t$, and using the bound on the average regret concludes the proof.
\end{proof}
\subsection{Fully Adaptive  Algorithm}
\label{sec:optimistism}
\begin{algorithm}[t]
\caption{Delay Optimistic Adaptive Anytime Online to Batch}
 \label{Alg:AnytimeOptimistic}
 \begin{algorithmic}
    \State {\bfseries Input:} \# of iterations $T$ , $\w_1 \in \K$, weights $\bc{\alpha_t}_{t \in [T]}$
    \State set $\g_{0-\tau_0}=0$ 
    \For{$t=1$ {\bfseries to} $T$}
    \State  send $\alpha_t\M_\delaytaut=\alpha_t\g_{t-1-\tau_{t-1}}$ to $\mathcal{A}$ as the $t$'th hint
    \State $\x_t \gets \frac{\sum_{i=1}^t\alpha_i\w_{i}}{\alpha_{1:t}}$
    \State get $\g_\delaytaut$ from worker
    \State define $f_t(\x)=\g_\delaytaut^\top \x$
    \State send $\alpha_t f_t(\x)$ to $\mathcal{A}$
    \State get $\w_{t+1}$ from $\mathcal{A}$
   \EndFor
   \State \textbf{return} $\x_T$
 \end{algorithmic}

\end{algorithm}

In this subsection we propose an \emph{optimal delay-adaptive, delay-robust algorithm for the delayed setting}, which implicitly adapts both the variance and smoothness.
 Although our scheme in Alg.~\ref{Alg:AnytimeTaut} enables to use  a variety of OCO algorithms, it does not necessarily obtains the optimal rates for the delayed setting. Moreover, the OCO algorithm $\A$ that we employ may require the knowledge of the problem parameters (like noise variance and smoothness) in order to ensure such optimal performance.

 Our algorithmic scheme is depicted in Alg.~ \ref{Alg:AnytimeOptimistic}.
 It is similar to the one we present in Alg.~\ref{Alg:AnytimeTaut}, only now we consider specialized OCO algorithms $\A$. Concretely, we assume that $\A$ is an optimistic and adaptive OCO method.
 Here  we limit ourselves to OCO algorithms that receive a sequence of linear losses $f_t(\x): = \g_t^\top \x$. \\
 \textbf{Optimism:} an optimistic OCO algorithm, \cite{rakhlin2013optimization}, is an algorithm that receives a ``hint" $\M_t$ prior to choosing $\w_t$.
 When the hints are good estimates of the loss gradient at round $t$, i.e., $\M_t\approx\g_t$, such algorithms are able to use these hints in order to obtain better regret guarantees.\\
 \textbf{Adaptive Optimistic Algorithm:} An adaptive optimistic OCO method is a method that upon receiving a sequence of linear losses $\{f_t(x): = \g_t^\top \x\}_{t\in[T]}$, and a sequence of hints $\{\M_t\}_{t\in[T]}$, enables to ensure a (weighted) regret bound of the following form, 
\begin{align}\label{eq:OptimisticAdaptiveGuarantee}
\Reg_T(\w^*) &:= \sum_{t=1}^T \alpha_t \g_t^\top(\w_t-\w^*) \leq O\left( D \sqrt{\sum_{T=1}^T \alpha_t^2 \Vert \M_t - \g_t \Vert^2}\right)~,
\end{align}
 where $\alpha_t$'s are predefined weight vectors. Note that the bound indeed improves when  $\M_t\approx\g_t$.
 In the context of Alg.~\ref{Alg:AnytimeOptimistic}, the hint vector prior to round $t$ is $\M_\delaytaut: =\g_{t-1-\tau_{t-1}}$, and $\A$ receives the following loss sequence $\{f_t(\x): = \g_{t-\tau_t}^\top \x\}$.
 Adaptive optimistic methods are developed e.g. in \cite{rakhlin2013optimization}, as well as in \citet{mohri2016accelerating}. 
 \begin{remark} \label{Remark:OptimisticOGD}
     The most natural example of an adaptive optimistic OCO method  is the following Optimistic OGD algorithm \cite{rakhlin2013optimization}, $\forall t\geq 2$,
     \begin{gather*}\label{eq:OptimisticOGDrAKHLIN}
         \w_{t} = \Pi_\K(\y_{t-1} - \eta_t \alpha_t \M_t)~\&~
         \y_{t} = \Pi_\K(\y_{t-1} - \eta_t \alpha_t \g_t) 
     \end{gather*}
    where, 
    $
    \eta_t: = D/\sqrt{1+\sum_{i=1}^{t-1}\alpha_i^2\|\g_i-\M_i\|^2}~,
    $ 
    and $\y_1=\w_1$ is an initial arbitrary point in $\K$.\\
    Observe that optimistic OGD  utilizes an additional sequence $\{\y_t\}_{t\in[T]}$:  whenever a hint $\M_t$ is received, we use it to take a step from $\y_{t-1}$ to compute the next decision point $\w_t$. Then, once we observe the true feedback $\g_t$, we use it to compute $\y_t$ by taking a gradient step from $\y_{t-1}$.
\end{remark}

The scheme in Alg.~ \ref{Alg:AnytimeOptimistic} is a combination of optimistic adaptive OCO method together with the anytime online to batch conversion scheme, that we apply to the delayed stochastic setting.
It was previously shown in \cite{kavis2019unixgrad} and \cite{cutkosky2019anytime} that employing optimistic and adaptive OCO methods in the context of stochastic optimization enables to adapt to problem parameters like noise variance and smoothness. Our next statement shows   that combining them within the anytime scheme enables to obtain a fully adaptive algorithm for the delayed setting.

\begin{theorem}
\label{theorem:optimisticAnytime}
Consider the delayed stochastic setting, and assume that $f:\K\mapsto \R$ is convex and $L$-smooth.
Then, using Alg.~\ref{Alg:AnytimeOptimistic} with $\alpha_t=t$ and an optimistic adaptive OCO method $\A$ that satisfies Eq.~\eqref{eq:OptimisticAdaptiveGuarantee}, ensures an \emph{optimal convergence rate of,}
\begin{align}
    \Expec\left[f(\x_T)-f(\w^*)\right] \leq&~~ O\left(\frac{L D^2(1+\sqrt{\sigma^2_\tau + \mu_\tau^2})}{T^{3/2}}\right)\nonumber+ O\left(\frac{L D^2\mu_\tau}{T}+\frac{\sigma D}{\sqrt{T}}\right)~.
\end{align}
Note that Alg.~\ref{Alg:AnytimeOptimistic} requires neither knowledge of $L,\sigma$ nor information regarding the delays.
\end{theorem}
The rate in Theorem~\ref{theorem:optimisticAnytime} matches the optimal rate for this setting. We actually obtain an improvement over \cite{arjevani2018tight,stich2019error} since our bound implies that we do not degrade compared to the non-delayed setting as long as 
$\mu_\tau \leq O(\sqrt{T})$, while their bound necessitates $\tau_{\max} \leq O(\sqrt{T})$.

\begin{proof}[Proof sketch of Theorem~\ref{theorem:optimisticAnytime}] Using the bound on $\Vert \x_t-\x_\delaytaut\Vert$ that we have previously shown, together with the bounds on the variance  and smoothness, enables us to prove the following, 
\begin{align*}
    \Expec[\Vert \M_\delaytaut - \g_\delaytaut \Vert^2] &\leq O\left(\frac{L^2D^2(1+\tau_t^2)}{t^2} + \sigma^2\right)~.
\end{align*}

Then, by using some known algebraic inequalities, we can bound $\Expec[R_T(\w^*)]$ with \\ $O\left(LD^2\sqrt{T}(1+ \sqrt{\sigma^2_\tau + \mu_\tau^2}) + D\sigma T^{3/2}\right)$. By combining this with Theorem~\ref{theo:GeneralAnytimeDelayTaut}, we receive the desired bound.
\end{proof}
\section{Delay Adaptive Method for the Strongly Convex Objectives}
\label{sec:strongConv}

In this section we present a delay-adaptive algorithm for the delayed setting, assuming that the objective is not only smooth, but also strongly-convex. 
Previous works on the delayed strongly-convex setting \cite{stich2019error, arjevani2018tight} require the knowledge of all relevant problem parameters including smoothness, strong-convexity, noise variance and maximal delay. Conversely, we only require the knowledge of the strong-convexity parameter. Moreover, our algorithm applies to constraint problems, which resolves an open problem of \cite{stich2019error}.

Prior to handling the delayed setting, we first develop an adaptive  SGD variant (Alg.~\ref{alg:OptOGD}) that applies to the standard  strongly-convex, smooth, and constrained setting (Section~\ref{sec:SC_STANDARD}).
Particularly, Alg.~\ref{alg:OptOGD} obtains a rate of $O(1/T^2+\sigma^2/T)$ without any prior knowledge except for the strong-convexity parameter. This is the first adaptive  algorithm for this setting  which does not require knowledge of the objective smoothness and noise variance, which might be of independent interest.
 
Then in Section~\ref{sec:SC_Delay} we introduce Alg.~\ref{alg:OptOGDDelay}, an adaptation of Alg.~\ref{alg:OptOGD} for the delayed setting, and prove convergence rate of $O((\sigma^2_\tau+\mu_\tau^2)/T^2+\sigma^2/T)$. Concretely, Alg.~\ref{alg:OptOGDDelay} implicitly accommodates changes in the delays and does not assume prior knowledge of $\sigma$ or $L$. Conversely to the general convex case, our method here is not based on an anytime online-to-batch conversion.


\subsection{Adaptive Optimistic Algorithm without Delays}
\label{sec:SC_STANDARD}
Here we discuss the standard strongly-convex and smooth constrained setting, and develop an  adaptive algorithm for this case, which only requires the knowledge of the strong-convexity parameter.
Note that this setting was already analyzed in \cite{joulani2020simpler} and \cite{zhang2019stochastic}, nevertheless, their methods require the knowledge of $L$ and $\sigma$ \footnote{Actually \cite{zhang2019stochastic} do not require $\sigma$ explicitly, nevertheless, they do not achieve a bound that captures the refined dependence on $\sigma$. It seems that in order to capture this dependence they should encode $\sigma$ inside their learning rate.}, similarly to the methods developed in \cite{stich2019error}. Thus, it seems like adapting these methods to the delayed setting  necessitates to equip them with the knowledge of strong-convexity and maximal delay, $H$ and $\tau_{\max}$, as is apparent from \cite{stich2019error}.

Consider the following update rule,
\begin{algorithm}[H]
\caption{Optimistic Strongly-Convex OGD  } \label{alg:OptOGD}
\begin{algorithmic}
    \State {\bfseries Input:} \# of iterations $T$ , $\x_0=\y_0 \in \K$, weights $\bc{\alpha_t}_{t \in [T]}$, Strong-convexity $H$
    \For{$t=1$ {\bfseries to} $T$}
    \State Define $\eta_t = \frac{8}{H\alpha_{1:t}}$
	\Indent
		\State $\x_t= \arg \min\limits_{\x \in \K} \alpha_t \M_t^\top\x +\frac{1}{2\eta_t}\|\x- \y_{t-1}\|^2$
		\State  $\y_t = \arg \min\limits_{\y \in \K} \alpha_t \g_t^\top\y + \frac{1}{2\eta_t}\|\y- \y_{t-1}\|^2 $
	\EndIndent
    \EndFor
    \State \textbf{return} $\xbar_{T} \propto \sum_{t=1}^T \alpha_t \x_t$
\end{algorithmic}
\end{algorithm}
where $\M_t = \gradf{ \x_{t-1} }+\xib_{t-1}$ and $\g_t = \gradf{ \x_t }+\xib_t$.

Alg.~\ref{alg:OptOGD} depicts a gradient weighting scheme of  optimistic OGD algorithm \cite{rakhlin2013optimization}, that we adapt  to the strongly convex case. 
Indeed, as proven in Appendix~\ref{sec:xy_ProjEquiv}, Alg.~\ref{alg:OptOGD} is equivalent to the optimistic OGD scheme that we described in Remark~\ref{Remark:OptimisticOGD}, yet with a learning rate which is adapted to the strongly convex case,
\begin{remark}
     The update rule in Alg. \ref{alg:OptOGD} is equivalent to,
     \begin{gather*}\label{eq:EquivSC}
         \x_{t} = \Pi_\K(\y_{t-1} - \teta_t \M_t)~~~\&~~~
         \y_{t} = \Pi_\K(\y_{t-1} - \teta_t \g_t) 
     \end{gather*}
    where
    $
    \teta_t: = \frac{8\alpha_t}{H\alpha_{1:t}}~.
    $
    \label{cor:xy_ProjEquiv}
\end{remark}

As we mention in Section~\ref{sec:optimistism}, optimistic methods can implicitly utilize smoothness to ensure better performance. When the objective is also strongly-convex, we can further improve our convergence guarantees as stated in Theorem ~\ref{thm:MainSC}  below.

\begin{theorem}
\label{thm:MainSC}
Assume that $f:\K\mapsto \R$ is $H$-strongly convex and $L$-smooth. Then, using Alg.~\ref{alg:OptOGD} with $\alpha_t =t^2$ ensures,
\begin{align*}
     \Expec[f(\xbar_{T}) - f(\w^*)] \leq& O\left( \frac{(G^2+\sigma^2)/H}{T^2} + \frac{\sigma^2}{TH}\right)  ~.
\end{align*}
\end{theorem}
The proof of Theorem~\ref{thm:MainSC} is described fully in Appendix~\ref{sec:Proof_thm:MainSC}.

\subsection{Delayed Constrained Setting}
\label{sec:SC_Delay}
Consider the following update rule when delays are present,
\begin{algorithm}[H]
\caption{Optimistic Strongly-Convex OGD with Delays} \label{alg:OptOGDDelay}
\begin{algorithmic}
    \State {\bfseries Input:} \# of iterations $T$ , $\x_0=\y_0 \in \K$, weights $\bc{\alpha_t}_{t \in [T]}$, Strong-convexity $H$
    \For{$t=1$ {\bfseries to} $T$}
    \State Define $\eta_t = \frac{8}{H\alpha_{1:t}}$
	 \Indent
        \State $\x_t= \arg \min\limits_{\x \in \K} \alpha_t \x^\top\M_\delaytaut +\frac{1}{2\eta_t}\|\x- \y_{t-1}\|^2 $
		\State $\y_t = \arg \min\limits_{\y \in \K} \alpha_t \y^\top\g_\delaytaut  + \frac{1}{2\eta_t}\|\y- \y_{t-1}\|^2  $
	 \EndIndent
    \EndFor
    \State \textbf{return} $\xbar_{T} \propto \sum_{t=1}^T \alpha_t \x_t$
\end{algorithmic}
\end{algorithm}
where $\M_\delaytaut = \gradf{ \x_\delaytautone }+\xib_\delaytautone$ and $\g_\delaytaut = \gradf{ \x_\delaytaut }+\xib_\delaytaut$.

The main statement for the strongly convex case under delayed feedback is stated below.
\begin{theorem}
\label{theorem:DelayedSC}
Assume that $f:\K\mapsto \R$ is $H$-strongly convex and $L$-smooth. Then using Alg.~\ref{alg:OptOGDDelay} with $\alpha_t =t^2$ ensures,
\begin{align*}
    \Expec[f(\xbar_{T}) - f(\w^*)] \leq&  O\left(\frac{(G^2+\sigma^2)}{HT^2}+\frac{\sigma^2}{HT}  \right)+  O\left(\frac{L^2(G^2+\sigma^2)(\sigma_\tau^2+\mu_\tau^2)}{H^3T^2} \right)~.
\end{align*}
\end{theorem}

Theorem~\ref{theorem:DelayedSC} offers a fully adaptive algorithm with convergence guarantees. Concretely, the slower converging term with variance dependency is similar to the one in \cite{stich2019error}, and does not depend upon the delays. 
The above bound implies that we do not degrade compared to non-delayed SGD as long as $\mu_\tau \leq O(\sqrt{T})$.

\begin{proof}[Proof Sketch (Theorem~\ref{theorem:DelayedSC})] 
Here we highlight the part in the proof that incorporates the delays, and demonstrate how we handle it. The additional term in the analysis that arises due to delays is the following,
\[\textrm{(D)} := \sum_{t=1}^{T} \alpha_t(\nabla f(\x_t)-\nabla f(\x_\delaytaut))^\top (\x_t-\w^*)~.\]

Using  $ab \leq \inf_{\rho>0}\left({\rho a^2}/{2}+{ b^2}/({2\rho})\right)$, together with the strong-convexity of $f(\cdot)$ which implies $H\|\x_t -\w^*\|^2\leq 2(f(\x_t)-f(\w^*))$ enables to bound $\textrm{(D)}$ as follows,
\begin{align*}
      \textrm{(D)}&\leq  \sum_{t=1}^{T}\frac{\alpha_tH}{8} \MyNorm{\x_t - \w^*}^2 +\frac{2\alpha_t}{H}\MyNorm{ \nabla f(\x_t)-\nabla f(\x_\delaytaut)}^2\\
      &\leq  \sum_{t=1}^{T}\frac{\alpha_t}{4} (f(\x_t)-f(\w^*))
      +\frac{2\alpha_t}{H}\MyNorm{ \nabla f(\x_t)-\nabla f(\x_\delaytaut)}^2
\end{align*}
The left term is directly related to the error, and the right term is then bounded using $\|\x_t -\x_\delaytaut\|\leq O(\tau_t/t)$ combined with the smoothness of the objective.
\end{proof}
\section{Experiments}
\label{sec:exp}

\begin{figure}[t]
\begin{center}
\centerline{\includegraphics[width=\columnwidth]{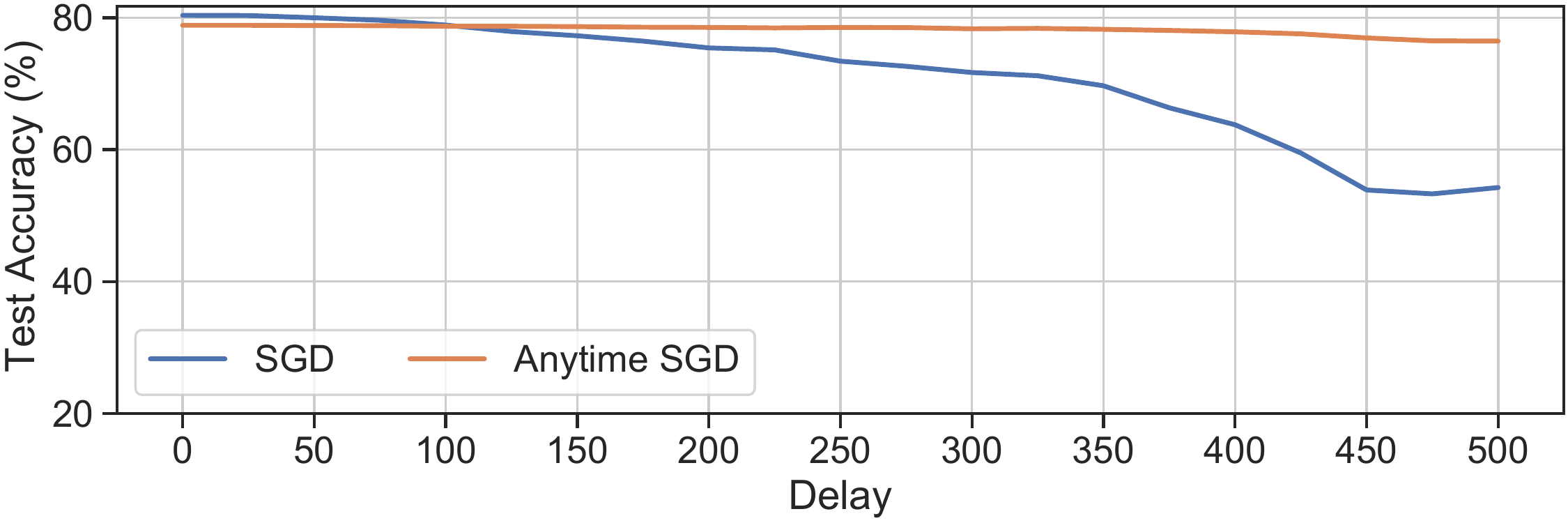}}
\caption{Accuracy as a function of update delays, with learning rate optimized for each of the algorithms for zero delay case.}
\label{fig:fashion_mnist_delay}
\end{center}
\end{figure}

\begin{figure}[t]
\begin{center} 
\centerline{\includegraphics[width=\columnwidth]{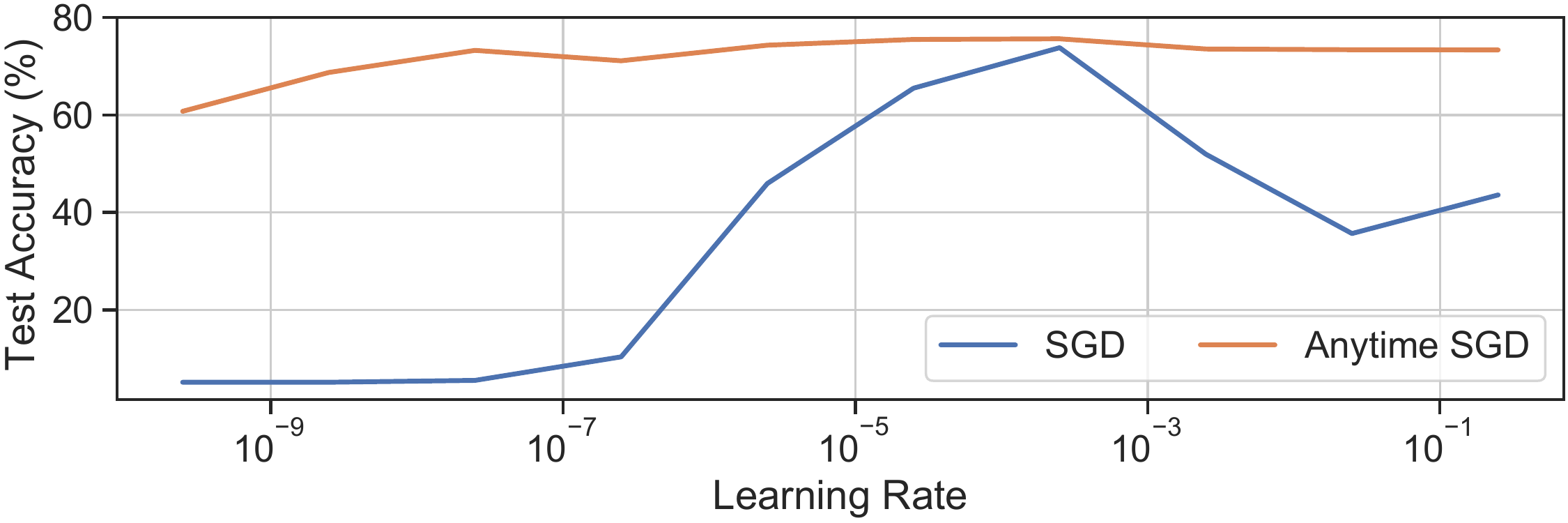}}
\caption{Accuracy as a function of learning rate
when $\tau_t=500$.}
\label{fig:fashion_mnist_robustness}
\end{center}
\end{figure}

In this section, we study the performance of anytime SGD: Algorithm~\ref{Alg:AnytimeTaut} using SGD as the OCO algorithm, as described in Eq.~\eqref{eq:SGDaDAPTIVEdEL}. We trained anytime SGD on Fashion-MNIST \citep{xiao2017fashion} dataset with logistic regression model and evaluated it using multi-class log loss. We compare our algorithm to SGD with validated constant step size, as was suggested in \cite{stich2019error}. Fashion-MNIST consists of $6\cdot10^4$ training examples and $1\cdot10^4$ test examples, when each example is a 28x28 grayscale clothing image, associated with a label from $10$ categories. The distributed system is simulated via an update queue, similarly to what was done in \cite{mcmahan2014delay}, which allows us to simulate delays of our choice. 

We show both scalability and robustness advantages of anytime SGD. As was mentioned before, on scalable systems the delay regime may change during the operation, making both these qualities essential for adequate performance. In addition, hyper-parameter tuning is expensive \citep{optuna_2019,bergstra2011algorithms,falkner2018bohb}, especially when required to tune with respect to different delay regimes.

Our scalability and delay adaptivity experiment is demonstrated in Figure~\ref{fig:fashion_mnist_delay}, where anytime SGD maintains high final accuracy even with significant amount of delay in the training. In this experiment we tuned the learning rate for each algorithm separately only in the zero delay setting, and then evaluate them on different delay regimes. Figure~\ref{fig:fashion_mnist_delay} shows that as the delay increases, the gap between the final accuracy of anytime SGD and SGD grows. Thus, while our algorithm maintains its high performance for different delay regimes with no further tuning, SGD greatly deteriorates. Note that similar results were received when tuning the algorithms with larger delay, and then changing the delay regime.

The robustness of anytime SGD is presented in Figure~\ref{fig:fashion_mnist_robustness}, where anytime SGD achieves high accuracy on a wide range of learning rates, whereas SGD achieves its top accuracy on a very narrow range. In this experiment we evaluated the algorithms with different learning rates, when $\tau_t=500$. This shows that anytime SGD is much less sensitive to learning changes than SGD.

For additional experimental results, please refer to Section~\ref{sec:ExpAppend} in the supplement.


\section{Discussion}
\label{sec:disc}
We leverage anytime online to batch scheme to prove convergence guarantees for delayed feedback setting, when updates are delayed to an unknown extent. We propose a delay adaptive training methods for constrained setting which do not depend on prior knowledge of problem parameters. We demonstrate experimentally (see Figures~\ref{fig:fashion_mnist_delay},\ref{fig:fashion_mnist_robustness}) that our algorithm outperforms alternative schemes for delayed feedback setting, exhibiting accuracy and robustness even when substantial delays are present. 

 One interesting direction for future research is to extend our approach to be used with accelerated first-order methods \cite{cutkosky2019anytime,kavis2019unixgrad}. 
 Moreover, our results suggest that incorporating anytime online to batch scheme has great potential for other challenging settings as well.  
 
 \section*{Acknowledgements}
 The work on this paper was supported in part by the Israeli Ministry of Science, Technology, and Space and by The Hasso Plattner Institute at the Technion. K.Y. Levy acknowledges support from the Israel Science Foundation (grant No. 447/20).

\bibliographystyle{abbrvnat}
\bibliography{Delaysbib}

\begin{thebibliography}{39}
\providecommand{\natexlab}[1]{#1}
\providecommand{\url}[1]{\texttt{#1}}
\expandafter\ifx\csname urlstyle\endcsname\relax
  \providecommand{\doi}[1]{doi: #1}\else
  \providecommand{\doi}{doi: \begingroup \urlstyle{rm}\Url}\fi

\bibitem[Agarwal and Duchi(2012)]{agarwal2012distributed}
A.~Agarwal and J.~C. Duchi.
\newblock Distributed delayed stochastic optimization.
\newblock In \emph{2012 IEEE 51st IEEE Conference on Decision and Control
  (CDC)}, pages 5451--5452. IEEE, 2012.

\bibitem[Akiba et~al.(2019)Akiba, Sano, Yanase, Ohta, and Koyama]{optuna_2019}
T.~Akiba, S.~Sano, T.~Yanase, T.~Ohta, and M.~Koyama.
\newblock Optuna: A next-generation hyperparameter optimization framework.
\newblock In \emph{Proceedings of the 25rd {ACM} {SIGKDD} International
  Conference on Knowledge Discovery and Data Mining}, 2019.

\bibitem[Arjevani et~al.(2020)Arjevani, Shamir, and Srebro]{arjevani2018tight}
Y.~Arjevani, O.~Shamir, and N.~Srebro.
\newblock A tight convergence analysis for stochastic gradient descent with
  delayed updates.
\newblock In \emph{Algorithmic Learning Theory}, pages 111--132. PMLR, 2020.

\bibitem[Ben-Nun and Hoefler(2019)]{ben2019demystifying}
T.~Ben-Nun and T.~Hoefler.
\newblock Demystifying parallel and distributed deep learning: An in-depth
  concurrency analysis.
\newblock \emph{ACM Computing Surveys (CSUR)}, 52\penalty0 (4):\penalty0 1--43,
  2019.

\bibitem[Bergstra et~al.(2011)Bergstra, Bardenet, Bengio, and
  K{\'e}gl]{bergstra2011algorithms}
J.~Bergstra, R.~Bardenet, Y.~Bengio, and B.~K{\'e}gl.
\newblock Algorithms for hyper-parameter optimization.
\newblock In \emph{25th annual conference on neural information processing
  systems (NIPS 2011)}, volume~24. Neural Information Processing Systems
  Foundation, 2011.

\bibitem[Bertsekas and Tsitsiklis(1989)]{bertsekas1989parallel}
D.~P. Bertsekas and J.~N. Tsitsiklis.
\newblock \emph{Parallel and distributed computation: numerical methods},
  volume~23.
\newblock Prentice hall Englewood Cliffs, NJ, 1989.

\bibitem[Cesa-Bianchi et~al.(2004)Cesa-Bianchi, Conconi, and
  Gentile]{cesa2004generalization}
N.~Cesa-Bianchi, A.~Conconi, and C.~Gentile.
\newblock On the generalization ability of on-line learning algorithms.
\newblock \emph{IEEE Transactions on Information Theory}, 50\penalty0
  (9):\penalty0 2050--2057, 2004.

\bibitem[Cotter et~al.(2011)Cotter, Shamir, Srebro, and
  Sridharan]{cotter2011better}
A.~Cotter, O.~Shamir, N.~Srebro, and K.~Sridharan.
\newblock Better mini-batch algorithms via accelerated gradient methods.
\newblock In \emph{Advances in neural information processing systems}, pages
  1647--1655, 2011.

\bibitem[Cutkosky(2019)]{cutkosky2019anytime}
A.~Cutkosky.
\newblock Anytime online-to-batch, optimism and acceleration.
\newblock In \emph{International Conference on Machine Learning}, pages
  1446--1454, 2019.

\bibitem[Dekel et~al.(2012)Dekel, Gilad-Bachrach, Shamir, and
  Xiao]{dekel2012optimal}
O.~Dekel, R.~Gilad-Bachrach, O.~Shamir, and L.~Xiao.
\newblock Optimal distributed online prediction using mini-batches.
\newblock \emph{Journal of Machine Learning Research}, 13\penalty0
  (Jan):\penalty0 165--202, 2012.

\bibitem[Dutta et~al.(2018)Dutta, Joshi, Ghosh, Dube, and
  Nagpurkar]{dutta2018slow}
S.~Dutta, G.~Joshi, S.~Ghosh, P.~Dube, and P.~Nagpurkar.
\newblock Slow and stale gradients can win the race: Error-runtime trade-offs
  in distributed sgd.
\newblock In \emph{International Conference on Artificial Intelligence and
  Statistics}, pages 803--812. PMLR, 2018.

\bibitem[Falkner et~al.(2018)Falkner, Klein, and Hutter]{falkner2018bohb}
S.~Falkner, A.~Klein, and F.~Hutter.
\newblock Bohb: Robust and efficient hyperparameter optimization at scale.
\newblock In \emph{International Conference on Machine Learning}, pages
  1437--1446. PMLR, 2018.

\bibitem[Feyzmahdavian et~al.(2016)Feyzmahdavian, Aytekin, and
  Johansson]{feyzmahdavian2016asynchronous}
H.~R. Feyzmahdavian, A.~Aytekin, and M.~Johansson.
\newblock An asynchronous mini-batch algorithm for regularized stochastic
  optimization.
\newblock \emph{IEEE Transactions on Automatic Control}, 61\penalty0
  (12):\penalty0 3740--3754, 2016.

\bibitem[Jain et~al.(2016)Jain, Kakade, Kidambi, Netrapalli, and
  Sidford]{jain2016parallelizing}
P.~Jain, S.~M. Kakade, R.~Kidambi, P.~Netrapalli, and A.~Sidford.
\newblock Parallelizing stochastic approximation through mini-batching and
  tail-averaging.
\newblock \emph{arXiv preprint arXiv:1610.03774}, 2016.

\bibitem[Joulani et~al.(2013)Joulani, Gyorgy, and
  Szepesv{\'a}ri]{joulani2013online}
P.~Joulani, A.~Gyorgy, and C.~Szepesv{\'a}ri.
\newblock Online learning under delayed feedback.
\newblock In \emph{International Conference on Machine Learning}, pages
  1453--1461. PMLR, 2013.

\bibitem[Joulani et~al.(2020)Joulani, Raj, Gyorgy, and
  Szepesvari]{joulani2020simpler}
P.~Joulani, A.~Raj, A.~Gyorgy, and C.~Szepesvari.
\newblock A simpler approach to accelerated optimization: iterative averaging
  meets optimism.
\newblock In \emph{International Conference on Machine Learning}, pages
  4984--4993. PMLR, 2020.

\bibitem[Kavis et~al.(2019)Kavis, Levy, Bach, and Cevher]{kavis2019unixgrad}
A.~Kavis, K.~Y. Levy, F.~Bach, and V.~Cevher.
\newblock Unixgrad: A universal, adaptive algorithm with optimal guarantees for
  constrained optimization.
\newblock In \emph{Advances in Neural Information Processing Systems}, pages
  6260--6269, 2019.

\bibitem[Leblond et~al.(2018)Leblond, Pedregosa, and
  Lacoste-Julien]{leblond2018improved}
R.~Leblond, F.~Pedregosa, and S.~Lacoste-Julien.
\newblock Improved asynchronous parallel optimization analysis for stochastic
  incremental methods.
\newblock \emph{arXiv preprint arXiv:1801.03749}, 2018.

\bibitem[Li et~al.(2014)Li, Zhang, Chen, and Smola]{li2014efficient}
M.~Li, T.~Zhang, Y.~Chen, and A.~J. Smola.
\newblock Efficient mini-batch training for stochastic optimization.
\newblock In \emph{Proceedings of the 20th ACM SIGKDD international conference
  on Knowledge discovery and data mining}, pages 661--670. ACM, 2014.

\bibitem[Lian et~al.(2015)Lian, Huang, Li, and Liu]{lian2015asynchronous}
X.~Lian, Y.~Huang, Y.~Li, and J.~Liu.
\newblock Asynchronous parallel stochastic gradient for nonconvex optimization.
\newblock \emph{arXiv preprint arXiv:1506.08272}, 2015.

\bibitem[McMahan and Streeter(2014)]{mcmahan2014delay}
H.~B. McMahan and M.~Streeter.
\newblock Delay-tolerant algorithms for asynchronous distributed online
  learning.
\newblock 2014.

\bibitem[Mohri and Yang(2016)]{mohri2016accelerating}
M.~Mohri and S.~Yang.
\newblock Accelerating online convex optimization via adaptive prediction.
\newblock In \emph{Artificial Intelligence and Statistics}, pages 848--856,
  2016.

\bibitem[Nedi{\'c} et~al.(2001)Nedi{\'c}, Bertsekas, and
  Borkar]{nedic2001distributed}
A.~Nedi{\'c}, D.~P. Bertsekas, and V.~S. Borkar.
\newblock Distributed asynchronous incremental subgradient methods.
\newblock \emph{Studies in Computational Mathematics}, 8\penalty0 (C):\penalty0
  381--407, 2001.

\bibitem[Needell et~al.(2013)Needell, Srebro, and Ward]{needell2013stochastic}
D.~Needell, N.~Srebro, and R.~Ward.
\newblock Stochastic gradient descent, weighted sampling, and the randomized
  kaczmarz algorithm.
\newblock \emph{arXiv preprint arXiv:1310.5715}, 2013.

\bibitem[Rakhlin and Sridharan(2013)]{rakhlin2013optimization}
A.~Rakhlin and K.~Sridharan.
\newblock Optimization, learning, and games with predictable sequences.
\newblock \emph{arXiv preprint arXiv:1311.1869}, 2013.

\bibitem[Recht et~al.(2011)Recht, Re, Wright, and Niu]{recht2011hogwild}
B.~Recht, C.~Re, S.~Wright, and F.~Niu.
\newblock Hogwild!: A lock-free approach to parallelizing stochastic gradient
  descent.
\newblock \emph{Advances in neural information processing systems},
  24:\penalty0 693--701, 2011.

\bibitem[Ren et~al.(2020)Ren, Zhou, Qiu, Deshpande, and
  Kalagnanam]{ren2020delay}
Z.~Ren, Z.~Zhou, L.~Qiu, A.~Deshpande, and J.~Kalagnanam.
\newblock Delay-adaptive distributed stochastic optimization.
\newblock In \emph{Proceedings of the AAAI Conference on Artificial
  Intelligence}, volume~34, pages 5503--5510, 2020.

\bibitem[Shalev-Shwartz and Zhang(2013)]{shalev2013accelerated}
S.~Shalev-Shwartz and T.~Zhang.
\newblock Accelerated mini-batch stochastic dual coordinate ascent.
\newblock In \emph{Advances in Neural Information Processing Systems}, pages
  378--385, 2013.

\bibitem[Shalev-Shwartz et~al.(2011)]{shalev2011online}
S.~Shalev-Shwartz et~al.
\newblock Online learning and online convex optimization.
\newblock \emph{Foundations and trends in Machine Learning}, 4\penalty0
  (2):\penalty0 107--194, 2011.

\bibitem[Shamir and Srebro(2014)]{shamir2014distributed}
O.~Shamir and N.~Srebro.
\newblock Distributed stochastic optimization and learning.
\newblock In \emph{2014 52nd Annual Allerton Conference on Communication,
  Control, and Computing (Allerton)}, pages 850--857. IEEE, 2014.

\bibitem[Sra et~al.(2015)Sra, Yu, Li, and Smola]{sra2015adadelay}
S.~Sra, A.~W. Yu, M.~Li, and A.~J. Smola.
\newblock Adadelay: Delay adaptive distributed stochastic convex optimization.
\newblock \emph{arXiv preprint arXiv:1508.05003}, 2015.

\bibitem[Stich and Karimireddy(2020)]{stich2019error}
S.~U. Stich and S.~P. Karimireddy.
\newblock The error-feedback framework: Better rates for sgd with delayed
  gradients and compressed updates.
\newblock \emph{Journal of Machine Learning Research}, 21:\penalty0 1--36,
  2020.

\bibitem[Tak{\'a}{\v{c}} et~al.(2015)Tak{\'a}{\v{c}}, Richt{\'a}rik, and
  Srebro]{takavc2015distributed}
M.~Tak{\'a}{\v{c}}, P.~Richt{\'a}rik, and N.~Srebro.
\newblock Distributed mini-batch sdca.
\newblock \emph{arXiv preprint arXiv:1507.08322}, 2015.

\bibitem[Verbraeken et~al.(2020)Verbraeken, Wolting, Katzy, Kloppenburg,
  Verbelen, and Rellermeyer]{verbraeken2020survey}
J.~Verbraeken, M.~Wolting, J.~Katzy, J.~Kloppenburg, T.~Verbelen, and J.~S.
  Rellermeyer.
\newblock A survey on distributed machine learning.
\newblock \emph{ACM Computing Surveys (CSUR)}, 53\penalty0 (2):\penalty0 1--33,
  2020.

\bibitem[Xiao et~al.(2017)Xiao, Rasul, and Vollgraf]{xiao2017fashion}
H.~Xiao, K.~Rasul, and R.~Vollgraf.
\newblock Fashion-mnist: a novel image dataset for benchmarking machine
  learning algorithms.
\newblock \emph{arXiv preprint arXiv:1708.07747}, 2017.

\bibitem[Zhang and Zhou(2019)]{zhang2019stochastic}
L.~Zhang and Z.-H. Zhou.
\newblock Stochastic approximation of smooth and strongly convex functions:
  Beyond the $ o (1/t) $ convergence rate.
\newblock In \emph{Conference on Learning Theory}, pages 3160--3179. PMLR,
  2019.

\bibitem[Zheng et~al.(2017)Zheng, Meng, Wang, Chen, Yu, Ma, and
  Liu]{zheng2017asynchronous}
S.~Zheng, Q.~Meng, T.~Wang, W.~Chen, N.~Yu, Z.-M. Ma, and T.-Y. Liu.
\newblock Asynchronous stochastic gradient descent with delay compensation.
\newblock In \emph{International Conference on Machine Learning}, pages
  4120--4129. PMLR, 2017.

\bibitem[Zhou et~al.(2018)Zhou, Mertikopoulos, Bambos, Glynn, Ye, Li, and
  Fei-Fei]{zhou2018distributed}
Z.~Zhou, P.~Mertikopoulos, N.~Bambos, P.~Glynn, Y.~Ye, L.-J. Li, and
  L.~Fei-Fei.
\newblock Distributed asynchronous optimization with unbounded delays: How slow
  can you go?
\newblock In \emph{International Conference on Machine Learning}, pages
  5970--5979. PMLR, 2018.

\bibitem[Zinkevich(2003)]{zinkevich2003online}
M.~Zinkevich.
\newblock Online convex programming and generalized infinitesimal gradient
  ascent.
\newblock In \emph{Proceedings of the 20th international conference on machine
  learning (icml-03)}, pages 928--936, 2003.

\end{thebibliography}

\newpage
\appendix
\section{Explaining the Model of Stochastic Delayed Oracle }
\label{sec:Noise}
In Section \ref{sec:Setting}, we model the delayed gradient oracle as follows,
$$
\g_\delaytaut = \nabla f(\x_\delaytaut) + \xib_t~,
$$
where $\Expec[\xib_t \vert\x_t]=0$, as well as $\Expec[\xib_t \vert\w_t]=0$

Recall that standard stochastic optimization problems in ML can be described as follows,
$$
\min_{\x\in\K}f(\x) : = \Expec_{z\sim \D}[f(\x;z)]~,
$$
when we assume we have an access to i.i.d.~samples 
$z_1,z_2\ldots z_T\sim \D$, which can be used to compute gradient estimates.
Thus, using this formulation, in the delayed setting we can assume that,
$$
\g_{t-\tau_t} = \nabla f(\x_{t-\tau_t};z_t)
$$
where $z_t$ is the random sample that is used by the (possibly stale)  machine that provides $\g_{t-\tau_t}$, which  is the gradient estimate that we use during our $t$'th update.

Since $z_1,z_2\ldots z_t$ are i.i.d., and since $\w_t$ and $\x_t$ depend only on $\w_1,z_1,z_2\ldots z_{t-1}$, which are independent of $z_t$, we have,

\begin{align}\label{eq:NoiseConditioned1}
    \Expec[\g_\delaytaut \vert \x_t] 
    =     \Expec \left[\Expec_{z_t}\left(\nabla f(\x_{\delaytaut};z_{t}) \vert \x_t,\x_{t-\tau_t}\right) \bigg\vert \x_t \right] =
     \Expec \left[\nabla f(\x_{\delaytaut})  \vert \x_t \right]~,
\end{align}
where we have used the law of total expectation.

Similarly,
\begin{align}\label{eq:NoiseConditioned2}
    \Expec[\g_\delaytaut \vert \w_t] 
    =     \Expec \left[\Expec_{z_t}\left(\nabla f(\x_{\delaytaut};z_{t}) \vert \w_t,\x_{t-\tau_t}\right) \bigg\vert \w_t \right] =
     \Expec \left[\nabla f(\x_{\delaytaut})  \vert \w_t \right]~.
\end{align}

Thus, the above Equations immediately imply $\Expec[\xib_t \vert\x_t]=0$, as well as $\Expec[\xib_t \vert\w_t]=0$.

\section{Proof of Theorem \ref{theo:GeneralAnytimeDelayTaut}} 
\begin{proof}
First, note that from the definition of $\x_t$,
$$
    \alpha_t(\x_t-\w_t)=\alpha_{1:t-1}(\x_{t-1}-\x_t)~,
$$
where we define $\alpha_{1:0}=0$ and $\x_0$ is an arbitrary element in $\K$. Now, we use the standard gradient inequality to obtain: 
\begin{align}
    \Expec\left[\sum_{t=1}^T \alpha_t(f(\x_t)-f(\w^*))\right]\leq&~ \Expec\left[\sum_{t=1}^T \alpha_t\nabla f(\x_t)^\top(\x_t-\w^*)\right]\nonumber\\
    =&~ \Expec\left[\sum_{t=1}^T \alpha_t\nabla f(\x_t)^\top(\x_t-\w_t)+\sum_{t=1}^T\alpha_t\nabla f(\x_t)^\top(\w_t-\w^*)\right]\nonumber\\
    =&~\Expec\left[\sum_{t=1}^T \alpha_{1:t-1}\nabla f(\x_t)^\top(\x_{t-1}-\x_t)+\sum_{t=1}^T\alpha_t\nabla f(\x_\delaytaut)^\top(\w_t-\w^*)\right]\nonumber\\
    &+\Expec\left[\alpha_t(\nabla f(\x_t)-\nabla f(\x_\delaytaut))^\top(\w_t-\w^*)\right]\nonumber\\
    \leq&~\Expec\left[\sum_{t=1}^T \alpha_{1:t-1}\nabla f(\x_t)^\top(\x_{t-1}-\x_t)+\sum_{t=1}^T\alpha_t\g_\delaytaut^\top(\w_t-\w^*)\right]\nonumber\\
    &+\Expec\left[\sum_{t=1}^T\alpha_t\Vert\nabla f(\x_t)-\nabla f(\x_\delaytaut)\Vert\Vert\w_t-\w^*\Vert\right]~,
    \label{eq:ExpBasicAnytime} 
\end{align}
where the third line uses the note above $\alpha_t(\x_t-\w_t)=\alpha_{1:t-1}(\x_{t-1}-\x_t)$, and the last is due to Cauchy-Schwarzt inequality and the definition of $\g_\delaytaut$.

The second term of Equation \eqref{eq:ExpBasicAnytime} is bounded by the OCO algorithm regret. Focusing on the last term, we wish to apply smoothness assumption to bound $\Vert\nabla f(\x_t)-\nabla f(\x_\delaytaut)\Vert$. For this purpose, we will examine the difference in iterate average $\Vert \x_t-\x_\delaytaut\Vert$, and show that $\Vert \x_t-\x_\delaytaut\Vert \leq O(\tau_t/t)$.

\paragraph{Bounding $\Vert \x_t-\x_\delaytaut\Vert$ :}

Let $\z$ be the tail average of $\x_t$ after $\delaytaut$, i.e.
$$
\z: = \frac{1}{\alpha_{\delaytaut+1:t}}\sum_{i=\delaytaut+1}^t \alpha_i \w_i~.
$$

Clearly $\z\in \K$ and the following holds,
\begin{align*}
    \alpha_{1:t} \x_t = \sum_{i=1}^t \alpha_i \w_i 
    =\sum_{i=1}^{\delaytaut} \alpha_i \w_i + \sum_{i=\delaytaut+1}^{t} \alpha_i \w_i
    = \alpha_{1:\delaytaut} \x_{\delaytaut} + \alpha_{\delaytaut+1:t}\z~.
    \end{align*}

Therefore, $ \alpha_{1:\delaytaut} (\x_t-\x_{\delaytaut}) = \alpha_{\delaytaut+1:t} (\z-\x_t)$, and by taking $\alpha_t=t$, we obtain:
\begin{align*}
    \|\x_t-\x_{\delaytaut}\| &= \frac{\alpha_{\delaytaut+1:t}}{\alpha_{1:\delaytaut}}\|\z-\x_t\|\\
    &=\frac{\tau_t(\delaytaut+1+t)}{(\delaytaut)(\delaytaut+1)}\|\z-\x_t\|\\
    &=\left[\frac{2\tau_t t}{(\delaytaut) (\delaytaut+1)}\right]\|\z-\x_t\|+\left[\frac{\tau_t(1-\tau_t)}{(\delaytaut)(\delaytaut+1)}\right]\|\z-\x_t\|\\
    &\leq\left[\frac{2\tau_t t}{(\delaytaut)^2}\right]\|\z-\x_t\|~.
\end{align*}

If $t\geq 2\tau_t\;$ we have:
\begin{align*}
    \|\x_t-\x_{\delaytaut}\| &\leq \frac{8\tau_t D}{t}~.
\end{align*}
Now Recall that the domain is bounded and therefore $ \|\x_t-\x_{\delaytaut}\|\leq D~;~\forall t$. In addition, for $t<2\tau_t$, we have $D<\frac{8\tau_t D}{t}$. Combining this with the above equation we conclude that,
\begin{align}
    \|\x_t-\x_{\delaytaut}\| &\leq \frac{8\tau_t D}{t}\qquad \forall t, \tau_t\leq t~.
    \label{eq:Boundx}
\end{align}

Using the property of smooth functions, 
\begin{equation}
    \Vert \nabla f(\x_t)-\nabla f(\x_\delaytaut)\Vert \leq L\Vert \x_t- \x_\delaytaut\Vert \leq \frac{8\tau_t LD}{t}~.
    \label{eq:GradDifBound}
\end{equation}

\paragraph{Final Bound :}

Combining Equations \eqref{eq:ExpBasicAnytime}, \eqref{eq:GradDifBound} together with $\alpha_t=t$ and $\|w_t-w^*\|\leq D$ yields:
\begin{align*}
     \Expec\left[\sum_{t=1}^T \alpha_t(f(\x_t)-f(\w^*))\right]\leq&
     \Expec\left[\sum_{t=1}^T \alpha_{1:t-1}\nabla f(\x_t)^\top(\x_{t-1}-\x_t)+\sum_{t=1}^T\alpha_t\g_\delaytaut^\top(\w_t-\w^*)\right]\\
    &+\Expec\left[\sum_{t=1}^T\alpha_t\Vert\nabla f(\x_t)-\nabla f(\x_\delaytaut)\Vert\Vert\w_t-\w^*\Vert\right]\\
     =&\Expec\left[\sum_{t=1}^T \alpha_{1:t-1}\nabla f(\x_t)^\top(\x_{t-1}-\x_t)\right]+\Expec[\Reg_T(\w^*)]+\sum_{t=1}^T 8\tau_t LD^2\\
     =&\Expec\left[\sum_{t=1}^T \alpha_{1:t-1}\nabla f(\x_t)^\top(\x_{t-1}-\x_t)\right]+\Expec[\Reg_T(\w^*)]+8LD^2T\mu_\tau~.
\end{align*}
where $\mu_\tau=\frac{\sum_{t=1}^T \tau_t}{T}$ is the average delay.

Next, we follow similar steps  as in the proof of  Theorem 1 of \cite{cutkosky2019anytime}. Using gradient inequality we have, $\Expec\left[ \nabla f(x_t)^\top(\x_{t-1}-\x_t)\right]\leq\Expec\left[f(\x_{t-1})-f(\x_t)\right]$. Hence,
\[\Expec\left[\sum_{t=1}^T \alpha_t(f(\x_t)-f(\w^*))\right]\leq \Expec\left[\sum_{t=1}^T \alpha_{1:t-1}(f(\x_{t-1})-f(\x_t))\right]+\Expec[\Reg_T(\w^*)]+8LD^2T\mu_\tau~.\]

By subtracting $\Expec\left[\sum_{t=1}^T \alpha_t f(\x_t)\right]$ from both sides of the equation we obtain,

\begin{gather*}
    -\alpha_{1:T} \Expec\left[f(\w^*))\right]\leq \Expec\left[\sum_{t=1}^T \alpha_{1:t-1}f(\x_{t-1})-\alpha_{1:t}f(\x_t)\right]+\Expec[\Reg_T(\w^*)]+8LD^2T\mu_\tau~.
\end{gather*}

Telescoping the above sum and dividing by $\alpha_{1:T}=\frac{T(T+1)}{2}$ conveys,
\begin{equation}
    \Expec\left[f(\x_T)-f(\w^*)\right] \leq \Expec\left[\frac{2\Reg_T(\w^*)}{T^2}\right] + \frac{16LD^2\mu_\tau}{T+1}~,
    \label{eq:AnytimeFinBound}
\end{equation}
as desired.
\end{proof}

\section{SGD for Delayed Setting}
\label{sec:DelayedSGD}

\begin{lemma}
Assume that $f:\K\mapsto \R$ is $L$-smooth. Let the online learning algorithm, $\mathcal{A}$, be SGD algorithm, with update rule
\[\w_{t+1} = \Pi_\K(\w_t-\eta_t \alpha_t\g_\delaytaut)~.\] 

Then, for $\alpha_t=1$, $\eta_t=\frac{D}{\sqrt{t(2G^2+2\sigma^2)}}$ we obtain, 
\[\Expec\left[\sum_{t=1}^{T}\alpha_t\g_\delaytaut^\top(\w_t-\w^*)\right] \leq \frac{3D\sqrt{T(2G^2+2\sigma^2)}}{2}~,\]
while for $\alpha_t=t$, $\eta_t=\frac{D}{\sqrt{t^3(2G^2+2\sigma^2)}}$ we obtain,
\[\Expec\left[\sum_{t=1}^{T}\alpha_t\g_\delaytaut^\top(\w_t-\w^*)\right] \leq D T^{3/2}\sqrt{2G^2+2\sigma^2}~.\]

The above bounds yield, \[O\left(\frac{\Reg_T(\w^*)}{\alpha_{1:T}} \right)=O\left(D\sqrt{\frac{G+\sigma}{T}}\right)~.\]
\end{lemma}
\textbf{Remark:} Note that in both cases if we denote the effective learning rate by $\teta_t = \alpha_t\eta_t$, implies $\teta_t =\theta(1/\sqrt{t})$.

\begin{proof}
    Using the Pythagorean Theorem we obtain,
    \begin{align*}
        \Vert \w_{t+1}-\w^* \Vert ^2 \leq&~ \Vert \w_t - \eta_t\alpha_t \g_\delaytaut-\w^* \Vert^2\nonumber\\
        =&~ \Vert \w_t - \w^* \Vert^2-2\eta_t\alpha_t \g_\delaytaut^\top (\w_t-\w^*)+\eta_t^2\alpha_t^2 \Vert  \g_\delaytaut\Vert^2~.
    \end{align*}

    Re-arranging, 
    \begin{align*}
        2\alpha_t\g_\delaytaut^\top (\w_t-\w^*) \leq&~ \frac{\Vert \w_{t}-\w^* \Vert ^2-\Vert \w_{t+1}-\w^* \Vert ^2}{\eta_t}+\eta_t\alpha_t^2 \Vert \g_\delaytaut\Vert^2\\
        \leq&~ \frac{\Vert \w_{t}-\w^* \Vert ^2-\Vert \w_{t+1}-\w^* \Vert ^2}{\eta_t}+\eta_t \alpha_t^2(2G^2+2\sigma^2)~.
    \end{align*}
    
    Summing from $t=1$ to $T$ in expectation and applying $\alpha_t=1$, $\eta_t=\frac{D}{\sqrt{t(2G^2+2\sigma^2)}}$ gives,
    \begin{align*}
        2\Expec\left[\sum_{t=1}^{T}\g_\delaytaut^\top(\w_t-\w^*)\right] &\leq D^2 \sum_{t=1}^{T}\left(\frac{1}{\eta_t}-\frac{1}{\eta_{t-1}}\right)+(2G^2+2\sigma^2)\sum_{t=1}^{T}\eta_t\nonumber\\
        & \leq \frac{D^2}{\eta_T} + (2G^2+2\sigma^2)\sum_{t=1}^{T}\eta_t\nonumber\\
        &\leq 3D\sqrt{T(2G^2+2\sigma^2)}~,
    \end{align*}
    where we define $\frac{1}{\eta_0}=0$. In the last line we used $\sum_{t=1}^T \frac{1}{\sqrt{t}}\leq 2\sqrt{T}$.
    
    Summing from $t=1$ to $T$ in expectation and applying $\alpha_t=t$, $\eta_t=\frac{D}{\sqrt{t^3(2G^2+2\sigma^2)}}$ gives,
    \begin{align*}
        2\Expec\left[\sum_{t=1}^{T}t\g_\delaytaut^\top(\w_t-\w^*)\right] &\leq D^2 \sum_{t=1}^{T}\left(\frac{1}{\eta_t}-\frac{1}{\eta_{t-1}}\right)+(2G^2+2\sigma^2) \sum_{t=1}^{T} t^2\eta_t\nonumber\\
        & \leq \frac{D^2}{\eta_T} +(2G^2+2\sigma^2) \sum_{t=1}^{T}t^2\eta_t \nonumber\\
        &\leq 2DT^{3/2}\sqrt{2G^2+2\sigma^2}~,
    \end{align*}
    where we again define $\frac{1}{\eta_0}=0$. In the last line we used $\sum_{t=1}^T \sqrt{t} \leq T^{3/2}$.
    
\end{proof}
\newpage
\section{Proof of theorem \ref{theorem:optimisticAnytime}}
\begin{proof}
As stated in Theorem \ref{theorem:optimisticAnytime}, under optimistic OCO algorithm, we assume,
\begin{align}
    \Reg_T(\w^*):= \sum_{t=1}^T \alpha_t \g_\delaytaut^\top(\w_t-\w^*) \leq O\left( D \sqrt{\sum_{T=1}^T \alpha_t^2 \Vert \M_\delaytaut - \g_\delaytaut \Vert^2}\right)
\end{align}

To bound the regret, we examine one summand, which depends on the difference between the hint $\M_\delaytaut = \g_\delaytautone$, and the received gradient $\g_\delaytaut$.
\begin{align*}
    \Vert \M_\delaytaut - \g_\delaytaut \Vert &= \Vert \g_\delaytaut- \g_\delaytautone \Vert \nonumber\\
    &\leq \Vert \nabla f(\x_\delaytaut)-\nabla f(\x_\delaytautone)\Vert+\Vert \xib_t-\xib_{t-1} \Vert\nonumber \\
    &\leq L\Vert \x_\delaytaut-\x_\delaytautone\Vert+\Vert \xib_t-\xib_{t-1} \Vert\nonumber \\
    &\leq L\Vert \x_t-\x_{t-1}\Vert+ L\Vert \x_{t-1}-\x_\delaytautone\Vert+ L\Vert \x_t-\x_\delaytaut\Vert+\Vert \xib_t\Vert+\Vert\xib_{t-1} \Vert\nonumber \\
    &\leq\frac{2LD}{t-1} + \frac{8LD\tau_{t-1}}{t-1}+\frac{8LD\tau_t}{t}+\Vert \xib_t\Vert+\Vert\xib_{t-1} \Vert ~,
\end{align*}
where the first inequality is achieved by plugging in the equation for the gradients and triangle inequality and the second uses smoothness. The third line uses again triangle inequality, and in the forth line we plugged in Eq.~\eqref{eq:Boundx}.

Using the Root Mean Square and Arithmetic Mean inequality, i.e. $\frac{\sum_{i=1}^n a_i}{n} \leq \sqrt{\frac{\sum_{i=1}^n a_i^2}{n}}$, we obtain:
\begin{align*}
    \Expec[\Vert \M_\delaytaut - \g_\delaytaut \Vert^2]&=\left(\frac{2LD}{t-1}+\frac{8LD\tau_{t-1}}{t-1}+\frac{8LD\tau_t}{t}+\Vert \xib_t\Vert+\Vert\xib_{t-1} \Vert\right)^2\\
    &\leq 5\left(\frac{4L^2D^2}{(t-1)^2}+\frac{64L^2D^2\tau_{t-1}^2}{(t-1)^2}+\frac{64L^2D^2\tau_t^2}{t^2}+2\sigma^2\right) ~.
\end{align*}

Therefore, for $\alpha_t=t$,
\begin{align*}
    \Expec&[\Reg_T(\w^*)] \leq \Expec\left[D \sqrt{2\sum_{t=1}^T \alpha_t^2 \Vert \M_\delaytaut - \g_\delaytaut \Vert^2 }\right]\\
    &\leq D \sqrt{\sum_{t=1}^T 160L^2D^2+\sum_{t=1}^T 2560L^2D^2\tau_{t-1}^2+640L^2D^2\sum_{t=1}^T\tau_t^2 + 20\sigma^2 \sum_{t=1}^Tt^2}\\
    &\leq D \sqrt{160 L^2D^2T+640L^2D^2\sum_{t=1}^T (4\tau_{t-1}^2+\tau_t^2) + 20\sigma^2\sum_{t=1}^T t^2}\\
    &\leq 13 LD^2\sqrt{T}+ D \sqrt{3200L^2D^2\sum_{t=1}^T \tau_t^2} + D\sqrt{20\sigma^2\sum_{t=1}^T t^2}\\
    &\leq 13 LD^2\sqrt{T}+57LD^2 \sqrt{T(\sigma^2_\tau + \mu_\tau^2)} + 5D\sigma(T+1)^{3/2}~,
\end{align*}
where $\mu_\tau$ is the delay average and $\sigma_\tau^2$ is the delay variance. The third line uses $\tau_0=0 $ to combine the sums, the forth line exploits the known inequality $\sqrt{\sum_{i=1}^N a_i} \leq \sum_{i=1}^N \sqrt{a_i}$, and in the last we plugged in $\sum_{t=1}^T t^2\leq\frac{3(T+1)^3}{2}$.

Adding this to Equation \eqref{eq:AnytimeFinBound}, concludes the proof:
\begin{align*}
    \Expec&\left[f(\x_T)-f(\w^*)\right]= O\left(\frac{L D^2(1+\sqrt{\sigma^2_\tau + \mu_\tau^2})}{T^{3/2}}+ \frac{\sigma D}{\sqrt{T}}+\frac{LD^2\mu_\tau}{T}\right)
\end{align*}
\end{proof}

\newpage
\section{Proof of Theorem ~\ref{thm:MainSC}}
\label{sec:Proof_thm:MainSC}

\begin{proof}
First we state a technical lemma that will be used throughout the proof of Theorem~\ref{thm:MainSC}. Its proof is given in Section \ref{sec:Proof_Alphas}.

\begin{lemma}
\label{lemma:Alphas}
For any $t\geq1$ let $\alpha_t =t^2$, and $\eta_t =C \frac{1}{\alpha_{1:t}}$ where $C$ is some constant. Also define $\alpha_0:=\alpha_1$ and $\eta_0: = \eta_1$.
Then, the following holds $\forall 0\leq t\leq s$,
$$
\alpha_t \eta_t \geq \alpha_s \eta_s~; ~~~\&~~~ \alpha_t^2 \eta_t \leq 4\alpha_{t-1}^2 \eta_{t-1}
$$
\end{lemma}

In addition, we require the following Lemma (see proof in Section~\ref{sec:Proof_lemma:OptimisticLinearSC}),

\begin{lemma}
\label{lemma:OptimisticLinearSC}
Under the same conditions of Theorem~\ref{thm:MainSC}, the following holds,
\begin{align} \label{eq:MainSC1}
			\sum_{t=1}^{T} \alpha_t (\x_t - \w^*)^\top\g_t 
		        ~\leq &~ \,  \underbrace{\sum_{t=1}^{T} \frac{\alpha_t^2 \eta_{t}}{2} \norm{\g_t - \M_t}^2}_{\textrm{(i)}} 
		        + 2 \sum_{t=1}^{T-1} \br{ \frac{1}{\eta_{t+1}} - \frac{1}{\eta_t} } \|\x_{t+1}-\w^*\|^2 
		          \nonumber\\
			 &
			+ 2\sum_{t=1}^{T-1} \br{ \frac{1}{\eta_{t+1}} - \frac{1}{\eta_t} } \norm{\x_{t+1} - \y_t}^2  + \frac{1}{\eta_1} D^2
	\end{align}
\end{lemma}
Next, we  relate term $\textrm{(i)}$ to $\sum_{t=1}^{T} \alpha_t (\x_t - \w^*)^\top\gradf{\x_t}$. To do so, we  require the following lemma,
\begin{lemma}\label{lem:CoCoercivity1}
Let $f:\K\mapsto \R$ be an $L$-smooth function, then,
$$
\|\nabla f(\x) -\nabla f(\y)\|^2 \leq L(\nabla f(\x) - \nabla f(\y))^\top(\x-\y)
$$ 
Concretely, if $\y \in\argmin_{x\in\K}f(\x)$ then for any $\x\in\K$ we have
$$
\|\nabla f(\x)-\nabla f(\y) \|^2 \leq L\nabla f(\x)^\top (\x-\y)
$$ 
\end{lemma}
\begin{proof}[Proof of Lemma~\ref{lem:CoCoercivity1}]
The first part is proven in \cite{needell2013stochastic}.
For the second part, notice that if $\y \in\argmin_{\x\in\K}f(\x)$ then optimality condition imply that $\nabla f(\y)^\top(\x-\y)\geq 0~;~\forall \x\in\K$. Combining this with the first part of the lemma establishes the second part.
\end{proof}

Using Lemma~\ref{lem:CoCoercivity1} we obtain,
\begin{align}
    \| \g_t-\M_t\|^2  =& \|\nabla f(\x_t) - \nabla f(\x_{t-1})+\xib_t-\xib_{t-1}\|^2 \nonumber\\
     =& \|\nabla f(\x_t) - \nabla f(\w^*) + \nabla f(\w^*) - \nabla f(\x_{t-1})+\xib_t-\xib_{t-1}\|^2 \nonumber\\
     \leq & 4\|\nabla f(\x_t)-\nabla f(\w^*)\|^2 + 4\|\nabla f(\x_{t-1})-\nabla f(\w^*)\|^2+4\|\xib_t\|^2+4\|\xib_{t-1}\|^2 \nonumber\\
     \leq&
     4L\nabla f(\x_t)^\top(\x_t-\w^*) + 4L\nabla f(\x_{t-1})^\top(\x_{t-1}-\w^*)+4\|\xib_t\|^2+4\|\xib_{t-1}\|^2
\end{align}
where we used $\frac{\sum_{i=1}^n a_i}{n} \leq \sqrt{\frac{\sum_{i=1}^n a_i^2}{n}}$ in the third line. 

Thus, we can bound $\textrm{(i)}$ as follows,
\begin{align}
\label{eq:Term1SC}
\textrm{(i)} : =& \sum_{t=1}^{T} \frac{\alpha_t^2 \eta_{t}}{2} \norm{\g_t - \M_t}^2\nonumber\\
\leq&
2L\sum_{t=1}^{T} \alpha_t^2 \eta_{t} \nabla f(\x_t)^\top(\x_t-\w^*) +2 L\sum_{t=1}^{T} \alpha_t^2 \eta_{t} \nabla f(\x_{t-1})^\top(\x_{t-1}-\w^*)\nonumber\\
&+2\sum_{t=1}^{T} \alpha_t^2 \eta_{t}(\|\xib_t\|^2+\|\xib_{t-1}\|^2)\nonumber\\
\leq&
2L\sum_{t=1}^{T} \alpha_t^2 \eta_{t} \nabla f(\x_t)^\top(\x_t-\w^*) + 8L\sum_{t=1}^{T} \alpha_{t-1}^2 \eta_{t-1} \nabla f(\x_{t-1})^\top(\x_{t-1}-\w^*)\nonumber\\
&+2\sum_{t=1}^{T}\|\xib_t\|^2 (\alpha_t^2 \eta_{t}+\alpha_{t+1}^2 \eta_{t+1})\nonumber\\
\leq&
10L \sum_{t=0}^T\alpha_t^2 \eta_{t} \nabla f(\x_t)^\top(\x_t-\w^*)  +10\sum_{t=1}^{T} \alpha_t^2 \eta_{t}\|\xib_t\|^2
\end{align}
where we define $\xib_0=0$. The third and forth line use $\nabla f(\x_t)^\top(\x_t-\w^*)\geq 0$, which holds due to convexity of $f$ and optimality of $\w^*$, together with Lemma~\ref{lemma:Alphas}.

Now, lets define, $t^*: = \min\{ t: 10L\alpha_t^2 \eta_{t} \leq \frac{1}{2}\alpha_t\}$. Note that according to  Lemma~\ref{lemma:Alphas}  $\alpha_t\eta_t$ is monotonic decreasing and therefore   $ \forall t\geq t^*;~10L\alpha_t^2 \eta_{t} \leq \frac{1}{2}\alpha_t$.
Using this together with Eq.~\eqref{eq:Term1SC}, as well as using the convexity of $f$ that implies $\nabla f(\x_t)^\top (\x_t-\w^*)\geq 0 ~;\forall t$ gives,
\begin{align}
\label{eq:Term1SC_secBound}
\textrm{(i)} 
&\leq
10L \sum_{t=0}^T\alpha_t^2 \eta_{t} \nabla f(\x_t)^\top(\x_t-x^*)+10\sum_{t=1}^{T} \alpha_t^2 \eta_{t}\|\xib_t\|^2 \nonumber  \\
&=
10L \sum_{t=0}^{t^*-1}\alpha_t^2 \eta_{t} \nabla f(\x_t)^\top(\x_t-\w^*) + 10L\sum_{t=t^*}^T \alpha_t^2 \eta_{t} \nabla f(\x_t)^\top(\x_t-\w^*)
+10\sum_{t=1}^{T} \alpha_t^2 \eta_{t}\|\xib_t\|^2 \nonumber  \\
&\leq
10L \sum_{t=0}^{t^*-1}\alpha_t^2 \eta_{t} \nabla f(\x_t)^\top(\x_t-\w^*) + \frac{1}{2}\sum_{t=t^*}^T \alpha_t  \nabla f(\x_t)^\top(\x_t-\w^*)
+10\sum_{t=1}^{T} \alpha_t^2 \eta_{t}\|\xib_t\|^2\nonumber  \\
&\leq
10L \sum_{t=0}^{t^*-1}\alpha_t^2 \eta_{t} \nabla f(\x_t)^\top(\x_t-\w^*) + \frac{1}{2}\sum_{t=1}^T \alpha_t  \nabla f(\x_t)^\top(\x_t-\w^*)+10\sum_{t=1}^{T} \alpha_t^2 \eta_{t}\|\xib_t\|^2
 ~,
\end{align}
where in the last line we use again the fact that $\nabla f(\x_t)^\top (\x_t-\w^*)\geq 0 ~;\forall t$. 

Plugging the above into Eq.~\eqref{eq:MainSC1} and re-arranging we obtain,
\begin{align*} 
    \sum_{t=1}^{T} \alpha_t (\x_t - \w^*)^\top\g_t  =& ~\sum_{t=1}^{T} \alpha_t
	(\x_t - \w^*)^\top\nabla f(\x_t) + \sum_{t=1}^{T} \alpha_t
	(\x_t - \w^*)^\top\xib_t \\
	~\leq &~ \,  10L \sum_{t=0}^{t^*-1}\alpha_t^2 \eta_{t} \nabla f(x_t)^\top(\x_t-\w^*) +  \frac{1}{2}\sum_{t=1}^{T} \alpha_t
	(\x_t - \w^*)^\top\nabla f(\x_t)\nonumber\\
	&+10\sum_{t=1}^{T} \alpha_t^2 \eta_{t}\|\xib_t\|^2 +2 \sum_{t=1}^{T-1} \br{ \frac{1}{\eta_{t+1}} - \frac{1}{\eta_t} } \|\x_{t+1}-\w^*\|^2\nonumber\\
	 & + 2\sum_{t=1}^{T-1} \br{ \frac{1}{\eta_{t+1}} - \frac{1}{\eta_t} } \norm{\x_{t+1}- \y_t}^2	 + \frac{1}{\eta_1} D^2~.
\end{align*}
which implies,
\begin{align*} 
    \frac{1}{2}\sum_{t=1}^{T} \alpha_t
	(\x_t - \w^*)^\top\nabla f(\x_t) \leq &~ \,  10L \sum_{t=0}^{t^*-1}\alpha_t^2 \eta_{t} \nabla f(x_t)^\top(\x_t-\w^*) +10\sum_{t=1}^{T} \alpha_t^2 \eta_{t}\|\xib_t\|^2  
	\nonumber\\
	& + 2 \sum_{t=1}^{T-1} \br{ \frac{1}{\eta_{t+1}} - \frac{1}{\eta_t} } \|\x_{t+1}-\w^*\|^2  + \frac{1}{\eta_1} D^2 \nonumber\\
	& + 2\sum_{t=1}^{T-1} \br{ \frac{1}{\eta_{t+1}} - \frac{1}{\eta_t} } \norm{\x_{t+1}- \y_t}^2  -\sum_{t=1}^{T} \alpha_t
	(\x_t - \w^*)^\top\xib_t~.
\end{align*}

Combining the above with the strong-convexity of $f(\cdot)$ implies
\begin{align}
\label{eq:SC_ABC}
    \sum_{t=1}^{T} \alpha_t( f(\x_t)-&f(\w^*)) 
    \leq  \,\sum_{t=1}^{T} \left( \alpha_t (\x_t - \w^*)^\top\nabla f(\x_t) -\frac{\alpha_tH}{2}\|\x_t-\w^*\|^2  \right)\nonumber\\
	\leq & \, \underbrace{20L \sum_{t=0}^{t^*-1}\alpha_t^2 \eta_{t} \nabla f(\x_t)^\top(\x_t-\w^*) }_\textrm{(A)} +20\sum_{t=1}^{T} \alpha_t^2 \eta_{t}\|\xib_t\|^2 \nonumber\\
	&+ 4 \underbrace{\sum_{t=1}^{T-1} \br{ \frac{1}{\eta_{t+1}} - \frac{1}{\eta_t} - \frac{H}{8}\alpha_{t+1}} \|\x_{t+1}-\w^*\|^2 }_\textrm{(B)} + 4 \underbrace{\sum_{t=1}^{T-1} \br{ \frac{1}{\eta_{t+1}} - \frac{1}{\eta_t} } \norm{\x_{t+1} - \y_t}^2}_\textrm{(C)}\nonumber\\
	&+ \frac{2}{\eta_1} D^2 -2\sum_{t=1}^{T} \alpha_t
	(\x_t - \w^*)^\top\xib_t~.
	\end{align}
	
Next we bound the above three terms.

\paragraph{Bounding \textrm{(A)}} Using the expression for $\eta_t$ and assumption 1,3 we have,
\begin{align*}
     \textrm{(A)}
     &: = 10L \sum_{t=0}^{t^*-1}\alpha_t^2 \eta_{t} \nabla f(\x_t)^\top(\x_t-\w^*)  \\
     &\leq \frac{80LGD}{H}\sum_{t=0}^{t^*-1}\frac{\alpha_t^2}{\alpha_{1:t}}  \\
     &\leq\frac{480 LGD}{H}\sum_{t=0}^{t^*-1}(t+1)\\
      &\leq\frac{240 LGD}{H} (t^*)^2  
\end{align*}
where in the two last lines we have used $\alpha_t = t^2$. Now, using the weights together with the definition of $t^*: = \min\{ t: 10L\alpha_t^2 \eta_{t} \leq \frac{1}{2}\alpha_t\}$ and the expression of $\eta_t$ implies,
$$
t^* \leq 160\frac{L}{H}
$$

Plugging this back to the above bound on $ \textrm{(A)}$ we finally obtain,
\begin{align} \label{eq:BoundrA}
 \textrm{(A)} \leq 
 2\cdot10^6 GD \left( \frac{L}{H} \right)^3 
\end{align}

\paragraph{Bounding \textrm{(B)}}
Recalling that $\eta_{t+1} := \frac{8}{H\alpha_{1:t+1}}$ immediately implies,
\begin{align}\label{eq:BoundrB}
 \textrm{(B)}: = \sum_{t=1}^{T-1} \br{ \frac{1}{\eta_{t+1}} - \frac{1}{\eta_t} - \frac{H}{8}\alpha_{t+1}} \|\x_{t+1}-\w^*\|^2  =  0
\end{align}

\paragraph{Bounding  \textrm{(C)}}
To bound term $ \textrm{(C)}$ we use Remark \ref{cor:xy_ProjEquiv}, in conjunction with the contraction property of the projection operator to obtain in expectation,
$$
\Expec\left[\|\x_{t+1}-\y_t\|\right] \leq \teta_{t+1} \Expec\left[\|\M_{t+1}\|\right] \leq \teta_{t+1}\sqrt{2G^2+2\sigma^2}~.
$$

The above enables to bound term $\textrm{(C)}$
\begin{align}
\label{eq:BoundrC}
     \Expec\left[\textrm{(C)}\right]& : =\Expec\left[\sum_{t=1}^{T-1} \br{ \frac{1}{\eta_{t+1}} - \frac{1}{\eta_t} } \norm{\x_{t+1} - \y_t}^2 \right] \nonumber\\
     &=\frac{H}{4}\Expec\left[\sum_{t=1}^{T-1} \alpha_{t+1} \norm{\x_{t+1} - \y_t}^2\right] \nonumber\\
     &\leq\frac{H(G^2+\sigma^2)}{2}\sum_{t=1}^{T-1} \alpha_{t+1} \teta_{t+1}^2 \nonumber\\
     &=\frac{32(G^2+\sigma^2)}{H} \sum_{t=1}^{T-1}  \alpha_{t+1} \left( \frac{\alpha_{t+1}}{\alpha_{1:t+1}}\right)^2 \nonumber\\
      &\leq\frac{1200 (G^2+\sigma^2)}{H} \sum_{t=1}^{T-1}  \frac{(t+1)^6}{(t(t+1) (2t+3))^2} \nonumber\\
       &\leq\frac{1200 (G^2+\sigma^2)}{H}T
\end{align}
where we used $\alpha_t = t^2$.

\paragraph{Final Bound} 
Combining the bounds in Equations~\eqref{eq:BoundrA}~\eqref{eq:BoundrB} and ~\eqref{eq:BoundrC} and assumption 2 into Eq.~\eqref{eq:SC_ABC} and taking expectation implies,

\begin{align}
\label{eq:SC_FINAL}
    \Expec\left[\sum_{t=1}^{T} \alpha_t( f(\x_t)-f(\w^*)) \right] \leq & \, O\left(GD\left( {L}/{H} \right)^3  + (G^2+\sigma^2)/H)T + HD^2 +(\sigma^2/H)T^2 \right)
\end{align}
Recalling $\xbar_{T} \propto \sum_{t=1}^T \alpha_t \x_t$ and using Jensen's inequality established the theorem.
\end{proof}

\subsection{Proof of Lemma \ref{lemma:Alphas}}
\label{sec:Proof_Alphas}

\begin{proof}[Proof of Lemma~\ref{lemma:Alphas}]
When $s>t$,
\begin{gather*}
    \alpha_t \eta_t \geq \alpha_s \eta_s\\
    \Updownarrow\\
    \frac{1}{\alpha_t \eta_t}\leq \frac{1}{\alpha_s \eta_s}\\
    \Updownarrow\\
    2t+3+\frac{1}{t} \leq 2s+3+\frac{1}{s}\\
    \Updownarrow\\
    2(t-s)\leq \frac{t-s}{ts}
    \Updownarrow\\
    2\geq \frac{1}{ts}
\end{gather*}
which is true $\forall t,s\geq 1$. For $s=t$, $\alpha_t \eta_t \geq \alpha_s \eta_s$ is trivially true.

The second part follows from
\begin{gather*}
    \alpha_t^2 \eta_t \leq 4\alpha_{t-1}^2 \eta_{t-1}\\
    \Updownarrow\\
    \frac{t^3}{(t+1)(2t+1)}\leq \frac{4(t-1)^3}{t(2t-1)} 
\end{gather*}
which is true for $t\geq 2$. For $t=1$,  the inequality is true by definition of $\alpha_1, \eta_1, \alpha_0, \eta_0$.

\end{proof}

\subsection{Proof of Remark \ref{cor:xy_ProjEquiv}}
\label{sec:xy_ProjEquiv}

\begin{proof}
    According to the update rule for $\x_t$ as stated in Alg.~\ref{alg:OptOGD},
\begin{align*}
    \x_t =&~ \arg \min\limits_{\x \in \K} \left[\alpha_t \x^\top\M_t+\frac{1}{2\eta_t}\Vert \x-\y_{t-1}\Vert^2\right]\nonumber\\
    =&~ \arg \min\limits_{\x \in \K} \left[\frac{1}{2\eta_t}\left( \Vert\x\Vert^2-2\x^\top(\y_{t-1}-\alpha_t\eta_t\M_t)+\Vert\y_{t-1}\Vert^2\right)\right]\nonumber\\
     =&~ \arg \min\limits_{\x \in \K} \left[\MyNorm{\x- (\y_{t-1}-\alpha_t\eta_t\M_t)}^2\right] \nonumber\\
     =&~ \Pi_\K\left(\y_{t-1}-\alpha_t\eta_t\M_t\right)
\end{align*}

Plugging in $\eta_t = \frac{8}{H\alpha_{1:t}}$ concludes the first part of the Remark. The proof of the second part is equivalent to the above, for update rule $\y_t = \arg \min\limits_{\y \in \K} \left[\alpha_t \y^\top\g_t+\frac{1}{2\eta_t}\Vert \y-\y_{t-1}\Vert^2\right]$.
\end{proof}

\subsection{Proof of Lemma~\ref{lemma:OptimisticLinearSC}} \label{sec:Proof_lemma:OptimisticLinearSC}

\begin{proof}
	\begin{align}
		\sum_{t=1}^{T} \alpha_t (\x_t - \w^*)^\top\g_t =& \sum_{t=1}^{T} \underbrace{ \alpha_t (\x_t - \y_t)^\top(\g_t - \M_t) }_\textrm{(A)} + \underbrace{ \alpha_t (\x_t - \y_t)^\top\M_t }_\textrm{(B)} + \underbrace{ \alpha_t (\y_t - \w^*)^\top\g_t}_\textrm{(C)}
	\label{eq:linearSC}
	\end{align}

For simplicity,	we will denote $D_{\mathcal R}(\x, \y):=\frac{1}{2}\Vert \x-\y\Vert^2$. Note that since $\frac{1}{2}\Vert \x-\y\Vert^2$ is the Bregman Divergence of $\mathcal{R}(\x)=\frac{1}{2}\Vert \x\Vert^2$, inequalities of Bregman Divergence hold true. Specifically, we make use of the three point property,
\[D_{\mathcal R}(\x, \y)+D_{\mathcal R}(\y, \z) = D_{\mathcal R}(\x, \z)+\nabla_\z D_{\mathcal R}(\z, \y)^\top(\x-\y)~.\]

	\paragraph{Bounding (A)}
	\begin{align*}
			\sum_{t=1}^{T} \alpha_t (\x_t - \y_t)^\top(\g_t - \M_t) \leq& \sum_{t=1}^{T} \alpha_t \norm{\g_t - \M_t} \norm{\x_t - \y_t} \quad \text{(Cauchy Schwartz Inequality)} \\
			\leq& \sum_{t=1}^{T} \frac{\rho}{2} \norm{\g_t - \M_t}^2 + \frac{\alpha_t^2}{2 \rho} \norm{\x_t - \y_t}^2
	\end{align*}
	where the last line is due to Young's Inequality, $ab \leq \inf_{\rho>0}\left({\rho a^2}/{2}+{ b^2}/({2\rho})\right)$.
	
	By setting $\rho = \alpha_t^2 \eta_{t}$, we get the following upper bound for term (A)
	\begin{align*}
			\sum_{t=1}^{T} \alpha_t (\x_t - \y_t)^\top(\g_t - \M_t) \leq \sum_{t=1}^{T} \frac{\alpha_t^2 \eta_{t}}{2} \norm{\g_t - \M_t}^2 + \frac{1}{2 \eta_{t}} \norm{\x_t - \y_t}^2
	\label{eq:bound_Aterm}
	\end{align*}	
	
	\paragraph{Bounding (B)}
	\begin{align}
			\sum_{t=1}^{T} \alpha_t (\x_t - \y_t)^\top\M_t \leq& \sum_{t=1}^{T} \frac{1}{\eta_t} \nabla_x D_{\mathcal R}(\x_t, \y_{t-1})^T (\y_t - \x_t) \quad (\text{Optimality for } \x_t) \\
			=& \sum_{t=1}^{T} \frac{1}{\eta_t} \br{ D_{\mathcal R}(\y_t, \y_{t-1}) - D_{\mathcal R}(\x_t, \y_{t-1}) - D_{\mathcal R}(\y_t, \x_t) }\quad (\text{Bregman Divergence})
	\label{eq:bound_Bterm}
	\end{align}

	\paragraph{Bounding (C)}
	\begin{align}
			\sum_{t=1}^{T} \alpha_t (\y_t - \w^*)^\top\g_t \leq& \sum_{t=1}^{T} \frac{1}{\eta_t} \nabla_x D_{\mathcal R}(\y_t, \y_{t-1})^T (\w^* - \y_t) \quad (\text{Optimality for } \y_t)\\
			=& \sum_{t=1}^{T} \frac{1}{\eta_t} \br{ D_{\mathcal R}(\w^*, \y_{t-1}) - D_{\mathcal R}(\y_t, \y_{t-1}) - D_{\mathcal R}(\w^*, \y_t) }\quad (\text{Bregman Divergence})
	\label{eq:bound_Cterm}
	\end{align}
	
	\paragraph{Final Bound}
	Combining the bounds in Equations~\eqref{eq:bound_Aterm}~\eqref{eq:bound_Bterm} and ~\eqref{eq:bound_Cterm} into Eq.~\eqref{eq:linearSC} implies,
	\begin{align} 
	\label{eq:MainSC2}
			\sum_{t=1}^{T} \alpha_t (\x_t - \w^*)^\top\g_t \leq& \,  \sum_{t=1}^{T} \frac{\alpha_t^2 \eta_{t}}{2} \norm{\g_t - \M_t}^2 + \frac{1}{2 \eta_{t}} \norm{\x_t - \y_t}^2  \nonumber\\
			&+ \frac{1}{\eta_t} \br{ D_{\mathcal R}(\w^*, \y_{t-1}) - D_{\mathcal R}(\w^*, \y_t) - D_{\mathcal R}(\x_t, \y_{t-1}) - D_{\mathcal R}(\y_t, \x_t) }  \nonumber\\
			= & \, \sum_{t=1}^{T} \frac{\alpha_t^2 \eta_{t}}{2} \norm{\g_t - \M_t}^2 + \frac{1}{2 \eta_{t}} \norm{\x_t - \y_t}^2 \nonumber\\
			&+ \frac{1}{\eta_t} \br{ D_{\mathcal R}(\w^*, \y_{t-1}) - D_{\mathcal R}(\w^*, \y_t) - \frac{1}{2} \br{ \norm{\x_t - \y_{t}}^2 + \norm{\x_t - \y_{t-1}}^2 } }  \nonumber\\
			\leq& \, \sum_{t=1}^{T} \frac{\alpha_t^2 \eta_{t}}{2} \norm{\g_t - \M_t}^2 + \sum_{t=1}^{T-1} \br{ \frac{1}{\eta_{t+1}} - \frac{1}{\eta_t} } D_{\mathcal R}(\w^*, \y_t) 
			+ \frac{1}{\eta_1} D^2 \nonumber\\
                        =& \, \sum_{t=1}^{T} \frac{\alpha_t^2 \eta_{t}}{2} \norm{\g_t - \M_t}^2 + \sum_{t=1}^{T-1} \br{ \frac{1}{\eta_{t+1}} - \frac{1}{\eta_t} } \|\y_t-\w^*\|^2   + \frac{1}{\eta_1} D^2 \nonumber\\
		        \leq & \,  \sum_{t=1}^{T} \frac{\alpha_t^2 \eta_{t}}{2} \norm{\g_t - \M_t}^2+ 2\sum_{t=1}^{T-1} \br{ \frac{1}{\eta_{t+1}} - \frac{1}{\eta_t} } \|\x_{t+1}-\w^*\|^2  \nonumber\\
			 &+ 2\sum_{t=1}^{T-1} \br{ \frac{1}{\eta_{t+1}} - \frac{1}{\eta_t} } \norm{\x_{t+1} - \y_t}^2  + \frac{1}{\eta_1} D^2
	\end{align}
where in the third line we used $D_{\mathcal R}(\x, \y):=\frac{1}{2}\Vert \x-\y\Vert^2$.

This establishes the lemma.
\end{proof}	

\newpage
\section{Proof of Theorem \ref{theorem:DelayedSC}}
\label{sec:Proof_DelayedSC}
Recall that under assumptions 1,2 from Section~\ref{sec:Setting}, we can show the following bound on the expected norm of the gradients, $\Expec\|\g_t\|\leq \tilde{G}=\sqrt{2G^2+2\sigma^2}$. 
Nevertheless, working with this in-expectation bound makes the proof a bit cumbersome. Therefore, to simplify the analysis, from now on we will assume that $\Vert\g\Vert\leq \Tilde{G}$ with probability $1$. Both assumptions lead to exactly the same in expectation guarantees as those we state in Theorem~\ref{theorem:DelayedSC}.
\begin{proof}
 
We first Note that Lemma \ref{lemma:OptimisticLinearSC} is true for the delayed setting as well,
\begin{align}
	\label{eq:MainSC1_delay2}
    \sum_{t=1}^{T} \alpha_t (\x_t - \w^*)^\top\g_\delaytaut\leq&~  \underbrace{\sum_{t=1}^{T} \frac{\alpha_t^2 \eta_{t}}{2} \norm{\g_\delaytaut - \M_\delaytaut}^2}_{\textrm{(i)}} 
    + 2 \sum_{t=1}^{T-1} \br{ \frac{1}{\eta_{t+1}} - \frac{1}{\eta_t} } \|\x_{t+1}-\w^*\|^2 \nonumber\\
	& + 2\sum_{t=1}^{T-1} \br{ \frac{1}{\eta_{t+1}} - \frac{1}{\eta_t} } \norm{\x_{t+1} - \y_t}^2  + \frac{1}{\eta_1} D^2
\end{align}

Next, we relate term (i) to $\sum_{t=1}^{T} \alpha_t (\x_t - \w^*)^\top\gradf{\x_t}$. 
\begin{align}
     \|\g_\delaytaut-\M_\delaytaut\|^2  =&~ \|\nabla f(\x_\delaytaut) - \nabla f(\x_\delaytautone)+\xib_\delaytaut-\xib_\delaytautone\|^2 \nonumber\\
     \leq&~  5\|\nabla f(\x_\delaytaut)-\nabla  f(\x_t)\|^2 + 5\|\nabla f(\x_t) - \nabla f(\x_{t-1})\|^2\nonumber\\
     &+5\|\nabla f(\x_{t-1}) - \nabla f(\x_\delaytautone)\|^2+5(\|\xib_\delaytaut\|^2+\|\xib_\delaytautone\|^2) \nonumber\\
     \leq&~ 5\|\nabla f(\x_t) - \nabla f(\x_{t-1})\|^2+5L^2\|\x_t-\x_\delaytaut\|^2 +5L^2\|\x_{t-1}-\x_\delaytautone\|^2\nonumber\\
     &+5(\|\xib_\delaytaut\|^2+\|\xib_\delaytautone\|^2)\nonumber\\
     \leq&~ 10\|\nabla f(\x_t) - \nabla f(\w^*)\|^2 + 10\|\nabla f(\x_{t-1}) -\nabla f(\w^*)\|^2+5L^2\|\x_t-\x_\delaytaut\|^2 \nonumber\\
     &+5L^2\|\x_{t-1}-\x_\delaytautone\|^2+5(\|\xib_\delaytaut\|^2+\|\xib_\delaytautone\|^2) \nonumber\\
      \leq&~ 10L\nabla f(\x_t)^\top (\x_t-\w^*)+ 10L\nabla f(\x_{t-1})^\top(\x_{t-1}-\w^*)+5L^2\|\x_t-\x_\delaytaut\|^2 \nonumber\\
     &+5L^2\|\x_{t-1}-\x_\delaytautone\|^2+5(\|\xib_\delaytaut\|^2+\|\xib_\delaytautone\|^2)
      \label{eq:summandGradDif2}
\end{align}
where we used $\frac{\sum_{i=1}^n a_i}{n} \leq \sqrt{\frac{\sum_{i=1}^n a_i^2}{n}}$ and smoothness. The last line is due to Lemma \ref{lem:CoCoercivity1}.

Next, we wish to bound $\Vert \x_t-\x_\delaytaut\Vert$. Using Remark \ref{cor:xy_ProjEquiv} with the contraction property of the projection operator, we obtain for $\alpha_t=t^2$,
\begin{align*}
    \Vert \y_t-\y_{t-1}\Vert \leq& \teta_t \Vert\g_t\Vert~ \leq\frac{24\Tilde{G}}{Ht}~,\\
     \| \x_t-\y_t\| \leq& \teta_t  \| \M_t-\g_t \| \leq \frac{48\Tilde{G}}{Ht}~,\\
      \| \x_\delaytaut-\y_\delaytaut\|\leq& \teta_\delaytaut \| \M_\delaytaut-\g_\delaytaut \| \leq \frac{48\Tilde{G}}{H(\delaytaut)}~.
\end{align*}

Combining all of the above, we obtain
\begin{align*}
     \|\x_t-\x_\delaytaut\|  \leq&~ \|\x_t -\y_t\| + \| \y_\delaytaut-\x_\delaytaut\| +\sum_{s=\delaytaut}^t\|\y_s-\y_{s-1}\|\nonumber\\
    \leq&~ \frac{48\Tilde{G}}{Ht}+\frac{48\Tilde{G}}{H(\delaytaut)}+\frac{24\Tilde{G}}{H} \sum_{s=\delaytaut}^t \frac{1}{s}\nonumber\\
    \leq&~ \frac{48\Tilde{G}}{Ht}+\frac{24\Tilde{G}(\tau_t+2)}{H(\delaytaut)}\nonumber\\
    \leq&~ \frac{48\Tilde{G}}{Ht}+\frac{48\Tilde{G}(\tau_t+1)}{Ht}\nonumber\\
    \leq&~ \frac{48\Tilde{G}(\tau_t+2)}{Ht}
\end{align*}
where in the forth line we assume $2\tau_t+2 \leq t$. When $t<2\tau_t+2$, we have, \[\frac{2G}{H}\leq\frac{24\Tilde{G}}{H}\leq \frac{48\Tilde{G}(\tau_t+2)}{Ht}~.\]
Recalling that for strongly convex functions we have,
\begin{equation}
    \Vert \x-\y\Vert\leq \frac{\Vert \gradf{\x}-\gradf{\y}\Vert}{H}\leq \frac{2G}{H}~,
    \label{eq:strongConvEquiv}
\end{equation}
which we also prove in Subsection~\ref{sec:StrongConvEquivProof} for completeness, we obtain,
\begin{equation}
    \|\x_t-\x_\delaytaut\| \leq \frac{48\Tilde{G}(\tau_t+2)}{Ht}\quad \forall t~.
    \label{eq:XDifBound2}
\end{equation}

Combining Equations~\eqref{eq:summandGradDif2} and~\eqref{eq:XDifBound2}, we obtain
\begin{align}
     \textrm{(i)}:=&~\sum_{t=1}^{T} \frac{\alpha_t^2 \eta_{t}}{2} \norm{\g_\delaytaut - \M_\delaytaut}^2\nonumber\\
     \leq&~ 5L\sum_{t=1}^{T} \alpha_t^2 \eta_{t}\nabla f(\x_t)^\top (\x_t-\w^*)+ 5L\sum_{t=1}^{T} \alpha_t^2 \eta_{t}\nabla f(\x_{t-1})^\top(\x_{t-1}-\w^*)\nonumber\\
     &+ 3L^2\sum_{t=1}^{T} \alpha_t^2 \eta_{t}\|\x_t-\x_\delaytaut\|^2 +3L^2\sum_{t=1}^{T} \alpha_t^2 \eta_{t}\|\x_{t-1}-\x_\delaytautone\|^2\nonumber\\
     &+3\sum_{t=1}^{T} \alpha_t^2 \eta_{t}(\|\xib_\delaytaut\|^2+\|\xib_\delaytautone\|^2)\nonumber\\
     \leq&~ 5L\sum_{t=1}^{T} \alpha_t^2 \eta_{t}\nabla f(\x_t)^\top (\x_t-\w^*)+ 20L\sum_{t=1}^{T} \alpha_{t-1}^2 \eta_{t-1}\nabla f(\x_{t-1})^\top(\x_{t-1}-\w^*)\nonumber\\
     &+ 12L^2\sum_{t=1}^{T} \frac{2\cdot24^3\Tilde{G}^2(\tau_t+2)^2}{H^3t} +3L^2\eta_1D+12L^2\sum_{t=2}^{T} \frac{2\cdot 24^3\Tilde{G}^2(\tau_{t-1}+2)^2}{H^3(t-1)}\nonumber\\
     &+3\sum_{t=1}^{T} \alpha_t^2 \eta_{t}(\|\xib_\delaytaut\|^2+\|\xib_\delaytautone\|^2)\nonumber\\
    \leq&~ 25L\sum_{t=1}^{T} \alpha_t^2 \eta_{t}\nabla f(\x_t)^\top (\x_t-\w^*)+ \frac{2\cdot24^4L^2\Tilde{G}^2T(\sigma_\tau^2+\mu_\tau^2+4\mu_\tau+4)}{H^3}+3L^2\eta_1D\nonumber\\
     &+3\sum_{t=1}^{T} \alpha_t^2 \eta_{t}(\|\xib_\delaytaut\|^2+\|\xib_\delaytautone\|^2)
    \label{eq:Term1_delay2}
\end{align}
where in the second inequality we used Lemma \ref{lemma:Alphas} with $\nabla f(\x_t)^\top(\x_t-\w^*)\geq 0$; the third inequality uses again the bound $\nabla f(\x_t)^\top(\x_t-\w^*)\geq 0$. 

Now, lets define, $t^*: = \min\{ t: 25L\alpha_t^2 \eta_{t} \leq \frac{1}{2}\alpha_t\}$. Similarly to the proof of Theorem ~\ref{thm:MainSC}, since $\alpha_t\eta_t$ is monotonic decreasing, $ \forall t\geq t^*;~25L\alpha_t^2 \eta_{t} \leq \frac{1}{2}\alpha_t$.
Using this in \eqref{eq:Term1_delay2} together with the convexity of $f$ that implies $\nabla f(\x_t)^\top (\x_t-\w^*)\geq 0 ~;\forall t$ gives,
\begin{align}
\label{eq:Term1SC_secBound_delay2}
    \textrm{(i)} \leq&~  25L\sum_{t=1}^{T} \alpha_t^2 \eta_{t}\nabla f(\x_t)^\top (\x_t-\w^*)+3L^2\eta_1D+ \frac{2\cdot24^4L^2\Tilde{G}^2T(\sigma_\tau^2+\mu_\tau^2+4\mu_\tau+4)}{H^3}\nonumber\\
    &+3\sum_{t=1}^{T} \alpha_t^2 \eta_{t}(\|\xib_\delaytaut\|^2+\|\xib_\delaytautone\|^2) \nonumber  \\
    =&~ 25L\sum_{t=1}^{t^*-1} \alpha_t^2 \eta_{t}\nabla f(\x_t)^\top (\x_t-\w^*)+  25L\sum_{t=t^*}^{T} \alpha_t^2 \eta_{t}\nabla f(\x_t)^\top (\x_t-\w^*)+3L^2\eta_1D\nonumber\\
    &+ \frac{2\cdot24^4L^2\Tilde{G}^2T(\sigma_\tau^2+\mu_\tau^2+4\mu_\tau+4)}{H^3}+3\sum_{t=1}^{T} \alpha_t^2 \eta_{t}(\|\xib_\delaytaut\|^2+\|\xib_\delaytautone\|^2) \nonumber  \\
    \leq& 25L\sum_{t=1}^{t^*-1} \alpha_t^2 \eta_{t}\nabla f(\x_t)^\top (\x_t-\w^*)+  \frac{1}{2}\sum_{t=t^*}^{T} \alpha_t\nabla f(\x_t)^\top (\x_t-\w^*)+3L^2\eta_1D\nonumber\\
    &+ \frac{2\cdot24^4L^2\Tilde{G}^2T(\sigma_\tau^2+\mu_\tau^2+4\mu_\tau+4)}{H^3}+3\sum_{t=1}^{T} \alpha_t^2 \eta_{t}(\|\xib_\delaytaut\|^2+\|\xib_\delaytautone\|^2) \nonumber  \\
    \leq& 25L\sum_{t=1}^{t^*-1} \alpha_t^2 \eta_{t}\nabla f(\x_t)^\top (\x_t-\w^*)+  \frac{1}{2}\sum_{t=1}^{T} \alpha_t\nabla f(\x_t)^\top (\x_t-\w^*)+3L^2\eta_1D\nonumber\\
    &+ \frac{2\cdot24^4L^2\Tilde{G}^2T(\sigma_\tau^2+\mu_\tau^2+4\mu_\tau+4)}{H^3}+3\sum_{t=1}^{T} \alpha_t^2 \eta_{t}(\|\xib_\delaytaut\|^2+\|\xib_\delaytautone\|^2)\nonumber\\
    =& 25L\sum_{t=1}^{t^*-1} \alpha_t^2 \eta_{t}\nabla f(\x_t)^\top (\x_t-\w^*)+ \frac{1}{2} \sum_{t=1}^{T} \alpha_t\nabla f(\x_\delaytaut)^\top (\x_t-\w^*)\nonumber\\
    &+\frac{1}{2}\sum_{t=1}^{T} \alpha_t(\gradf{\x_t}-\nabla f(\x_\delaytaut))^\top (\x_t-\w^*)+3L^2\eta_1D\nonumber\\
    &+ \frac{2\cdot24^4L^2\Tilde{G}^2T(\sigma_\tau^2+\mu_\tau^2+4\mu_\tau+4)}{H^3}+3\sum_{t=1}^{T} \alpha_t^2 \eta_{t}(\|\xib_\delaytaut\|^2+\|\xib_\delaytautone\|^2)~,
\end{align}
where in the third inequality we use again the fact that $\nabla f(\x_t)^\top (\x_t-\w^*)\geq 0 ~;\forall t$. 

Plugging the above into Eq.~\eqref{eq:MainSC1_delay2} and re-arranging we obtain,
\begin{align*} 
    \sum_{t=1}^{T} \alpha_t (\x_t - \w^*)^\top&\g_\delaytaut = \sum_{t=1}^{T} \alpha_t (\x_t - \w^*)^\top\gradf{\x_\delaytaut} + \sum_{t=1}^{T} \alpha_t (\x_t - \w^*)^\top\xib_t\\
    \leq &~ \,  25L \sum_{t=0}^{t^*-1}\alpha_t^2 \eta_{t} \gradf{\x_t}^\top(\x_t-\w^*) + 2 \sum_{t=1}^{T-1} \br{ \frac{1}{\eta_{t+1}} - \frac{1}{\eta_t} } \|\x_{t+1}-\w^*\|^2 
	\nonumber\\
	&+ 2\sum_{t=1}^{T-1} \br{ \frac{1}{\eta_{t+1}} - \frac{1}{\eta_t} } \norm{\x_{t+1}- \y_t}^2 + \frac{1}{2}\sum_{t=1}^{T} \alpha_t\nabla f(\x_\delaytaut)^\top (\x_t-\w^*) \nonumber\\
	&+ \frac{1}{2}\sum_{t=1}^{T} \alpha_t(\gradf{\x_t}-\nabla f(\x_\delaytaut))^\top (\x_t-\w^*)+3L^2\eta_1D + \frac{1}{\eta_1} D^2\nonumber\\
	&+ \frac{2\cdot24^4L^2\Tilde{G}^2T(\sigma_\tau^2+\mu_\tau^2+4\mu_\tau+4)}{H^3}+3\sum_{t=1}^{T} \alpha_t^2 \eta_{t}(\|\xib_\delaytaut\|^2+\|\xib_\delaytautone\|^2)  ~,
\end{align*}
which implies,
\begin{align*}
    \frac{1}{2}\sum_{t=1}^{T} \alpha_t (\x_t - \w^*)^\top&\gradf{\x_\delaytaut}\leq ~ \,  25L \sum_{t=0}^{t^*-1}\alpha_t^2 \eta_{t} \gradf{\x_t}^\top(\x_t-\w^*) 
	\nonumber\\
	&+ 2\sum_{t=1}^{T-1} \br{ \frac{1}{\eta_{t+1}} - \frac{1}{\eta_t} } \norm{\x_{t+1}- \y_t}^2 + 2 \sum_{t=1}^{T-1} \br{ \frac{1}{\eta_{t+1}} - \frac{1}{\eta_t} } \|\x_{t+1}-\w^*\|^2  \nonumber\\
	&+ \frac{1}{2}\sum_{t=1}^{T} \alpha_t(\gradf{\x_t}-\nabla f(\x_\delaytaut))^\top (\x_t-\w^*)+3L^2\eta_1D + \frac{1}{\eta_1} D^2\nonumber\\
	&+ \frac{2\cdot24^4L^2\Tilde{G}^2T(\sigma_\tau^2+\mu_\tau^2+4\mu_\tau+4)}{H^3}+3\sum_{t=1}^{T} \alpha_t^2 \eta_{t}(\|\xib_\delaytaut\|^2+\|\xib_\delaytautone\|^2)\\
	& + \frac{1}{2}\sum_{t=1}^{T} \alpha_t\nabla f(\x_\delaytaut)^\top (\x_t-\w^*)-\sum_{t=1}^{T} \alpha_t (\x_t - \w^*)^\top\xib_t  ~.
\end{align*}
Combining the above with the strong-convexity of $f\cdot)$ implies,
\begin{align}
\label{eq:SC_ABC_Delay2}
    \sum_{t=1}^{T} \alpha_t( f(\x_t)-&f(\w^*))
    \leq  \,\sum_{t=1}^{T} \left( \alpha_t (\x_t - \w^*)^\top\nabla f(\x_t) -\frac{\alpha_tH}{2}\|\x_t-\w^*\|^2  \right)\nonumber\\
    = & \,\sum_{t=1}^{T} \left( \alpha_t (\x_t - \w^*)^\top\nabla f(\x_\delaytaut) + \alpha_t(\gradf{\x_t}-\nabla f(\x_\delaytaut))^\top (\x_t-\w^*)   \right)\nonumber\\
    &-\sum_{t=1}^{T}\frac{\alpha_tH}{2}\|\x_t-\w^*\|^2\nonumber\\
	\leq & \, \underbrace{50L \sum_{t=0}^{t^*-1}\alpha_t^2 \eta_{t} \gradf{\x_t}^\top(\x_t-\w^*)}_\textrm{(A)} + 4 \underbrace{\sum_{t=1}^{T-1} \br{ \frac{1}{\eta_{t+1}} - \frac{1}{\eta_t} - \frac{H}{8}\alpha_{t+1}} \|\x_{t+1}-\w^*\|^2 }_\textrm{(B)}\nonumber\\
	&+ 4 \underbrace{\sum_{t=1}^{T-1} \br{ \frac{1}{\eta_{t+1}} - \frac{1}{\eta_t} } \norm{\x_{t+1} - \y_t}^2}_\textrm{(C)}+ 2\underbrace{\sum_{t=1}^T \alpha_t (\gradf{\x_t}-\nabla f(\x_\delaytaut))^\top (\x_t - \w^*)}_\textrm{(D)}\nonumber\\
	&+ \frac{4\cdot24^4L^2\Tilde{G}^2T(\sigma_\tau^2+\mu_\tau^2+4\mu_\tau+4)}{H^3}+ \frac{2}{\eta_1} D^2+6L^2\eta_1D\nonumber\\
	&+6\sum_{t=1}^{T} \alpha_t^2 \eta_{t}(\|\xib_\delaytaut\|^2+\|\xib_\delaytautone\|^2)-2\sum_{t=1}^{T} \alpha_t (\x_t - \w^*)^\top\xib_t
	\end{align}
	
Terms $\textrm{(A)}-\textrm{(C)}$ can be bounded exactly as in Theorem \ref{thm:MainSC} proof, with $t^* \leq 320 \frac{L}{H}$, i.e.:
\begin{align*}
 \textrm{(A)} &\leq O\left( GD \left( \frac{L}{H} \right)^3\right)~, \\
  \textrm{(B)} &= 0~,\\
  \textrm{(C)} &= O\left(\frac{G^2T}{H}\right)~.
\end{align*}

So, we are left with bounding term $\textrm{(D)}$.
\paragraph{Bounding (D)}
\begin{align}
   \textrm{(D)}:=&\sum_{t=1}^{T}\alpha_t (\nabla f(\x_t)-\nabla f(\x_\delaytaut))^\top (\x_t - \w^*)\nonumber\\
    \leq&~\sum_{t=1}^{T}\alpha_t \MyNorm{\nabla f(\x_t)-\nabla f(\x_\delaytaut)}\MyNorm{\x_t - \w^*}\nonumber \\
    \leq & \sum_{t=1}^{T}\frac{\rho}{2}\MyNorm{\x_t - \w^*}^2  +\frac{\alpha_t^2}{2\rho}\MyNorm{ \nabla f(\x_t)-\nabla f(\x_\delaytaut)}^2\nonumber\\
    \leq & \sum_{t=1}^{T}\frac{\alpha_tH}{8} \MyNorm{\x_t - \w^*}^2 +\frac{2\alpha_t}{H}\MyNorm{ \nabla f(\x_t)-\nabla f(\x_\delaytaut)}^2\nonumber\\
    \leq & \sum_{t=1}^{T}\frac{\alpha_t}{4} (f(\x_t)-f(\w^*)) +\frac{2\alpha_tL}{H}\MyNorm{\x_t - \x_\delaytaut}^2 \nonumber\\
    \leq & \sum_{t=1}^{T}\frac{\alpha_t}{4} (f(\x_t)-f(\w^*)) +\frac{10^3L^2\Tilde{G}^2T(\sigma^2_\tau + \mu_\tau^2+4\mu_\tau+4)}{H^3}~,
    \label{eq:BoundrD2}
\end{align}
where the second line is due to Cauchy-Schwartz Inequality, the third line is due to Young's Inequality, $ab \leq \inf_{\rho>0}\left({\rho a^2}/{2}+{ b^2}/({2\rho})\right)$, when in the fourth line we took $\rho = \frac{\alpha_tH}{4}$. The fifth line utilizes smoothness and strong convexity which implies $\frac{H}{2}\|\x-\w^*\|^2\leq f(\x)-f(\w^*)$. The last line uses \eqref{eq:XDifBound2} and Root Mean Square with Arithmetic Mean inequality, i.e. $\frac{\sum_{i=1}^n a_i}{n} \leq \sqrt{\frac{\sum_{i=1}^n a_i^2}{n}}$. 

\paragraph{Final Bound} 
As we mentioned at the beginning of this proof, under assumption 1,2 from Section~\ref{sec:Setting}, in expectation we have $\Tilde{G}=\sqrt{2G^2+2\sigma^2}$. Combining the bounds in Equations~\eqref{eq:BoundrA}~\eqref{eq:BoundrB}~\eqref{eq:BoundrC}, ~\eqref{eq:BoundrD2} and assumption 2 into Eq.~\eqref{eq:SC_ABC_Delay2} and taking expectation implies,

\begin{align}
\label{eq:SC_FINAL_delay2}
    \Expec\left[\sum_{t=1}^{T} \alpha_t( f(\x_t)-f(\w^*)) \right] \leq & \, O\left(\frac{GDL^3}{H^3} + \frac{(G^2+\sigma^2)T}{H} + HD^2+\frac{L^2D^2}{H}\right)\nonumber\\
    &+O\left(\frac{L^2(G^2+\sigma^2)T(\sigma_\tau^2+\mu_\tau^2)}{H^3}+\frac{\sigma^2T^2}{H} \right)
\end{align}
Recalling $\xbar_{T} \propto \sum_{t=1}^T \alpha_t \x_t$ and using Jensen's inequality established the theorem.
\end{proof}

\subsection{Proof of Inequality~\eqref{eq:strongConvEquiv}}
\label{sec:StrongConvEquivProof}
\begin{proof}
Let us define $F(\x) \triangleq f(\x)-\frac{H}{2}\|\x\|^2$. Note that $F(\x)$ is convex, which follows from,
\begin{align*}
    F(\y)-F(\x) =&~ f(\y)-f(\x) -\frac{H}{2}(\|\y\|^2-\|\x\|^2)\\
    \geq&~ \gradf{\x}^\top(\y-\x) +\frac{H}{2}\|\x-\y\|^2-\frac{H}{2}(\|\y\|^2-\|\x\|^2)\\
    =&~ \gradf{\x}^\top(\y-\x)-H\x^\top(\y-\x)\\
    =&~ \nabla F(\x)^\top(\y-\x)~.
\end{align*}

From the monotone gradient condition for convexity of $F(\x)$ we obtain,
\begin{align*}
    (\gradf{\y}-\gradf{\x})^\top(\y-\x) =& (\nabla F(\y)-\nabla F(\x))^\top(\y-\x)+H(\y-\x)^\top(\y-\x)\geq H\|\y-\x\|^2\\~,
\end{align*}
where the second line uses strong convexity of $f(\cdot)$.
Using Cauchy-Schwartz on the above inequality gives,
\begin{align*}
    \|\gradf{\y}-\gradf{\x}\|\|\y-\x\|\geq (\gradf{\y}-\gradf{\x})^\top(\y-\x)\geq H\|\y-\x\|^2~.
\end{align*}

Dividing both sides of the inequality above by $\|\y-\x\|$ concludes the proof.
\end{proof}
\newpage
\section{Further Experiments}
\label{sec:ExpAppend}

\begin{figure}[t]
\begin{center}
\centerline{\includegraphics[width=0.9\columnwidth]{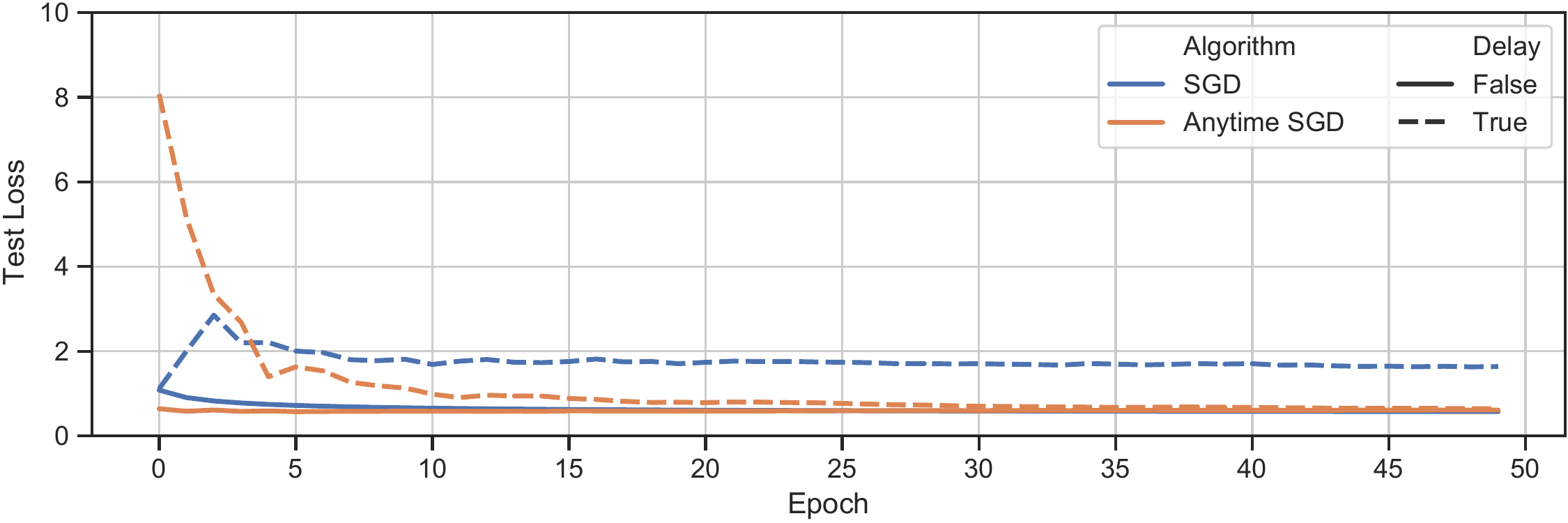}}
\caption{Expected excess loss as function of epochs when $\tau_t=500$, with learning rate optimized for each of the algorithms for zero delay regime.}
\label{fig:compare_stich}
\end{center}
\end{figure}

As was mentioned in Section \ref{sec:exp}, Figure \ref{fig:fashion_mnist_delay} demonstrates the final test accuracy for different delay regimes. Figure \ref{fig:compare_stich} expands on the regime of $\tau_t=500$, and compares between the expected excess loss of our algorithm and that of SGD as was suggested by \cite{stich2019error}. While the addition of the delay affects the convergence, as evident from the theoretical bounds as well, in anytime SGD the expected loss approaches the optimal one, while that of SGD does not.

\begin{figure}[t]
\begin{center}
\centerline{\includegraphics[width=0.9\columnwidth]{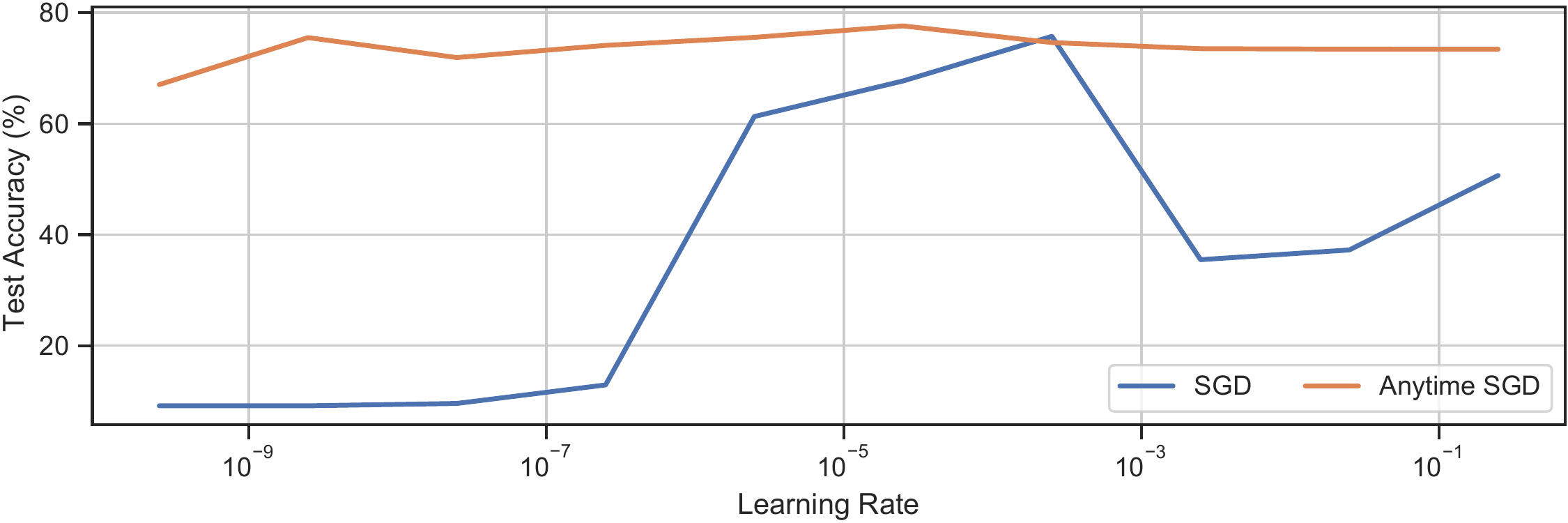}}
\caption{Accuracy as a function of learning rate
when $\tau_t\sim\text{Lognormal}(7, 0.4^2)$}
\label{fig:robustness_fattail}
\end{center}
\end{figure}

While Figure~\ref{fig:fashion_mnist_robustness} demonstrates the performance of the algorithms on a wide range of learning rates when $\tau_t=500$, in Figure~\ref{fig:robustness_fattail} the delay is distributed $\text{Lognormal}(7,0.4^2)$, a heavy-tail distribution. This shows that anytime SGD performs better when a high maximal delay, but reasonable mean, is present.

\end{document}